# Calibrated Multivariate Regression with Application to Neural Semantic Basis Discovery[*]

**Han Liu**                                                               HANLIU@PRINCETON.EDU
*Department of Operations Research and Financial Engineering,*
*Princeton University, NJ 08544, USA*

**Lie Wang**                                                             LIEWANG@MATH.MIT.EDU
*Department of Mathematics, Massachusetts Institute of Technology,*
*Cambridge MA 02139, USA*

**Tuo Zhao**[†]                                                            TOUR@CS.JHU.EDU
*Department of Computer Science, Johns Hopkins University,*
*Baltimore, MD 21218, USA*

**Editor:** Arthur Gretton

## Abstract

We propose a calibrated multivariate regression method named CMR for fitting high dimensional multivariate regression models. Compared with existing methods, CMR calibrates regularization for each regression task with respect to its noise level so that it simultaneously attains improved finite-sample performance and tuning insensitiveness. Theoretically, we provide sufficient conditions under which CMR achieves the optimal rate of convergence in parameter estimation. Computationally, we propose an efficient smoothed proximal gradient algorithm with a worst-case numerical rate of convergence $\mathcal{O}(1/\epsilon)$, where $\epsilon$ is a pre-specified accuracy of the objective function value. We conduct thorough numerical simulations to illustrate that CMR consistently outperforms other high dimensional multivariate regression methods. We also apply CMR to solve a brain activity prediction problem and find that it is as competitive as a handcrafted model created by human experts. The R package `camel` implementing the proposed method is available on the Comprehensive R Archive Network http://cran.r-project.org/web/packages/camel/.

## 1. Introduction

This paper studies the multivariate regression problem. Let $\mathbf{X} \in \mathbb{R}^{n \times d}$ be the design matrix and $\mathbf{Y} \in \mathbb{R}^{n \times m}$ be the response matrix, we consider a linear model

$$\mathbf{Y} = \mathbf{X}\mathbf{B}^0 + \mathbf{Z}, \tag{1}$$

where $\mathbf{B}^0 \in \mathbb{R}^{d \times m}$ is an unknown regression coefficient matrix and $\mathbf{Z} \in \mathbb{R}^{n \times m}$ is a noise matrix (Anderson, 1958; Breiman and Friedman, 2002). For a matrix $\mathbf{A} = [\mathbf{A}_{jk}] \in \mathbb{R}^{d \times m}$, we denote its $j^{\text{th}}$ row and $k^{\text{th}}$ column by $\mathbf{A}_{j*} = (\mathbf{A}_{j1}, ..., \mathbf{A}_{jm}) \in \mathbb{R}^m$ and $\mathbf{A}_{*k} = (\mathbf{A}_{1k}, ..., \mathbf{A}_{dk})^T \in \mathbb{R}^d$ respectively. We assume that all $\mathbf{Z}_{i*}$'s are independently sampled from an $m$-dimensional distribution with mean $\mathbf{0}$ and covariance matrix $\mathbf{\Sigma} \in \mathbb{R}^{m \times m}$.

*. Some preliminaries results in this paper were presented at the 28-th Annual Conference on Neural Information Processing Systems, Montreal, Quebec, Canada, 2014 (Liu et al., 2014a). This work is partially supported by grants NIH R01MH102339, NSF IIS1408910, NSF IIS1332109, NSF CAREER DMS1454377, NIH R01GM083084, NIH R01HG06841, and NSF Grant DMS-1005539.

†. Tuo Zhao is also affiliated with Department of Operations Research and Financial Engineering at Princeton University.

We can represent (1) as an ensemble of univariate linear regression models:

$$\mathbf{Y}_{*k} = \mathbf{X}\mathbf{B}^0_{*k} + \mathbf{Z}_{*k}, \ k = 1, ..., m,$$

which results in a multi-task learning problem (Baxter, 2000; Caruana, 1997; Caruana et al., 1996; Thrun, 1996; Ando and Zhang, 2005; Johnson and Zhang, 2008; Zhang et al., 2006; Zhang, 2006). Multi-task learning exploits shared common structure across tasks to obtain improved estimation performance. In the past decade, significant progress has been made on designing various modeling assumptions for multivariate regression.

One popular approach is to assume that the regression coefficients across different tasks are coupled by some shared common factors so that $\mathbf{B}^0$ has a low rank structure, i.e., $\text{rank}(\mathbf{B}^0) \ll \min(d, m)$. Under this assumption, a consistent estimator of $\mathbf{B}^0$ can be obtained by adopting either a non-convex rank constraint (Anderson, 1958; Izenman, 1975; Reinsel and Velu, 1998; Anderson, 1999; Reinsel and Velu, 1998; Izenman, 2008) or a convex relaxation using the nuclear norm regularization (Yuan et al., 2007; Amit et al., 2007; Argyriou et al., 2008; Negahban and Wainwright, 2011; Rohde and Tsybakov, 2011; Bunea et al., 2011, 2012; Bunea and Barbu, 2009; Mukherjee et al., 2012; Giraud, 2011; Argyriou et al., 2010; Foygel and Srebro, 2011; Johnson and Zhang, 2008; Salakhutdinov and Srebro, 2010; Evgeniou et al., 2006; Heskes, 2000; Teh et al., 2005; Yu et al., 2005). Such a low rank multivariate regression method is often applied to scenarios where $m$ is large.

Another approach is to assume that all the regression tasks share a common sparsity pattern, i.e., many $\mathbf{B}^0_{j*}$'s are zero vectors. Such a joint sparsity assumption for multivariate regressions is a natural extension from sparse univariate linear regressions. Similar to using the $L_1$-regularization in Lasso (Tibshirani, 1996; Chen et al., 1998), group regularization can be used to obtain a consistent estimator of $\mathbf{B}^0$ (Yuan and Lin, 2005; Turlach et al., 2005; Meier et al., 2008; Lounici et al., 2011; Kolar et al., 2011). Such a sparse multivariate regression method is often applied to scenarios where the dimension $d$ is large.

In this paper, we consider an uncorrelated structure for the noise matrix $\mathbf{Z}$, i.e.,

$$\mathbf{\Sigma} = \text{diag}(\sigma_1^2, \sigma_2^2, \dots, \sigma_{m-1}^2, \sigma_m^2). \tag{2}$$

Such an assumption allows us to efficiently solve the resulting estimation problem with a convex program and prove that the obtained estimator achieves the minimax optimal rates of convergence in parameter estimation.[1] For example, many existing work propose to solve the convex program

$$\widehat{\mathbf{B}} = \underset{\mathbf{B}}{\text{argmin}} \ \frac{1}{\sqrt{n}} ||\mathbf{Y} - \mathbf{X}\mathbf{B}||_{\text{F}}^2 + \lambda \mathcal{R}(\mathbf{B}), \tag{3}$$

where $\lambda > 0$ is a tuning parameter, $\mathcal{R}(\mathbf{B})$ is a regularization function of $\mathbf{B}$, and $||\mathbf{A}||_{\text{F}} = \sqrt{\sum_{j,k} \mathbf{A}_{jk}^2}$ is the Frobenius norm of a matrix $\mathbf{A}$. Popular choices of $\mathcal{R}(\mathbf{B})$ include

$$\text{Nuclear Norm} : ||\mathbf{B}||_* = \sum_{j=1}^{r} \psi_j(\mathbf{B}), \tag{4}$$

$$L_{1,p} \text{ Norm} : ||\mathbf{B}||_{1,p} = \sum_{j=1}^{d} \left( \sum_{k=1}^{m} |\mathbf{B}_{jk}|^p \right)^{1/p} \quad \text{for } 2 \leq p < \infty, \tag{5}$$

$$L_{1,\infty} \text{ Norm} : ||\mathbf{B}||_{1,\infty} = \sum_{j=1}^{d} \max_{1 \leq k \leq m} |\mathbf{B}_{jk}|, \tag{6}$$

---

1. See more details on exploiting the covariance structure of the noise matrix $\mathbf{Z}$ for multivariate regression in Breiman and Friedman (2002); Reinsel (2003); Rothman et al. (2010).

where $r$ in (4) is the rank of $\mathbf{B}$ and $\psi_j(\mathbf{B})$ represents the $j^{\text{th}}$ largest singular value of $\mathbf{B}$. The optimization problem (3) can be efficiently solved by the block coordinate descent algorithm (Liu et al., 2009a,b; Liu and Ye, 2010; Zhao et al., 2014a,c), fast proximal gradient algorithm (Toh and Yun, 2010; Beck and Teboulle, 2009a,b), and alternating direction method of multipliers(Boyd et al., 2011; Liu et al., 2014b). Scalable software packages such as MALSAR have been developed (Zhou et al., 2012).

The problem in (2) is amenable to statistical analysis. Under suitable conditions on the noise and design matrices, let $\sigma_{\max} = \max_k \sigma_k$ and $||\mathbf{X}||_2 = \psi_1(\mathbf{X})$ denote the largest singular value of $\mathbf{X}$, if we choose

$$\text{Low Rank}: \ \lambda = 2c \cdot \frac{||\mathbf{X}||_2}{n} \cdot \sigma_{\max} \left( \sqrt{d} + \sqrt{m} \right), \tag{7}$$

$$\text{Joint Sparsity}: \ \lambda = 2c \cdot \sigma_{\max} \left( \sqrt{\log d} + m^{1-1/p} \right), \tag{8}$$

for some $c > 1$, then the estimator $\widehat{\mathbf{B}}$ in (3) achieves the optimal rates of convergence[2] (Lounici et al., 2011; Rohde and Tsybakov, 2011). More specifically, there exists some universal constant $C$ such that, with high probability,

$$\text{Low Rank}: \ \frac{1}{\sqrt{m}}||\widehat{\mathbf{B}} - \mathbf{B}^0||_{\text{F}} \leq C \cdot \frac{||\mathbf{X}||_2}{\sqrt{n}} \cdot \sigma_{\max} \left( \sqrt{\frac{r}{n}} + \sqrt{\frac{rd}{nm}} \right),$$

$$\text{Joint Sparsity}: \ \frac{1}{\sqrt{m}}||\widehat{\mathbf{B}} - \mathbf{B}^0||_{\text{F}} \leq C \cdot \sigma_{\max} \left( \sqrt{\frac{s \log d}{nm}} + \sqrt{\frac{sm^{1-2/p}}{n}} \right),$$

where $r$ is the rank of $\mathbf{B}^0$ for the low rank setting and $s$ is the number of rows with non-zero entries in $\mathbf{B}^0$ for the setting of joint sparsity.

The estimator in (3) has two drawbacks: (i) All the tasks are regularized by the same tuning parameter $\lambda$, even though different tasks may have different $\sigma_k$'s. Thus more estimation bias is introduced to the tasks with smaller $\sigma_k$'s since they have to compensate the tasks with larger $\sigma_k$'s. In another word, these tasks are not calibrated (Zhao and Liu, 2014). (ii) The tuning parameter selection, as shown in (7) and (8), involves the unknown parameter $\sigma_{\max}$. This requires the regularization parameter to be carefully tuned over a wide range of potential values in order to get a good finite-sample performance.

To overcome the above two drawbacks, we propose a new method named calibrated multivariate regression (CMR) based on the convex program

$$\widehat{\mathbf{B}} = \underset{\mathbf{B}}{\text{argmin}} \, ||\mathbf{Y} - \mathbf{X}\mathbf{B}||_{2,1} + \lambda \mathcal{R}(\mathbf{B}) \tag{9}$$

where $||\mathbf{A}||_{2,1} = \sum_k \sqrt{\sum_j \mathbf{A}_{jk}^2}$ is the $L_{2,1}$ norm of a matrix $\mathbf{A} = [\mathbf{A}_{jk}] \in \mathbb{R}^{d \times m}$. This is a multivariate extension of the square-root Lasso estimator (Belloni et al., 2011; Sun and Zhang, 2012). Similar to the square-root Lasso, the tuning parameter selection of CMR does not involve $\sigma_{\max}$. Thus the resulting procedure adapts to different $\sigma_k$'s and achieves an improved finite-sample performance comparing with the ordinary multivariate regression estimator (OMR) defined in (3). Since both the loss and regularization functions in (9) are nonsmooth, CMR is computationally more challenging than OMR. To efficiently solve CMR, we develop a smoothed proximal gradient algorithm with a worst-case iteration complexity of $\mathcal{O}(1/\epsilon)$, where $\epsilon$ is a pre-specified accuracy of the objective value (Nesterov, 2005; Chen et al., 2012; Zhao and Liu, 2012; Zhao et al., 2014b). Theoretically, we show that under suitable conditions, CMR achieves the optimal rates of convergence in parameter

2. For the joint sparsity setting, the rate of convergence is optimal when $\mathbf{R}(\mathbf{B}) = ||\mathbf{B}||_{1,2}$. See more details in Lounici et al. (2011)

estimation. Numerical experiments on both synthetic and real data show that CMR universally outperforms existing multivariate regression methods. For a brain activity prediction task, prediction based on the features selected by CMR significantly outperforms that based on the features selected by OMR, and is even competitive with that based on the handcrafted features selected by human experts.

This paper is organized as follows: In §2, we describe the CMR method. In §3, we investigate the statistical properties of CMR; In §4, we derive a smoothed proximal gradient algorithm for solving CMR optimization. In §5, we conduct numerical experiments to illustrate the usefulness of the proposed method. In §6, we discuss the relationships between our results and other related work.

**Notation:** Given a vector $\boldsymbol{v} = (v_1, \ldots, v_d)^T \in \mathbb{R}^d$, for $1 \leq p \leq \infty$, we define the vector norms: $||\boldsymbol{v}||_p = \left( \sum_{j=1}^d |v_j|^p \right)^{1/p}$ for $1 \leq p < \infty$ and $||\boldsymbol{v}||_\infty = \max_{1 \leq j \leq d} |v_j|$. Given two matrices $\mathbf{A} = [\mathbf{A}_{jk}]$ and $\mathbf{C} = [\mathbf{C}_{jk}] \in \mathbb{R}^{d \times m}$, we define the inner product of $\mathbf{A}$ and $\mathbf{C}$ as $\langle \mathbf{A}, \mathbf{C} \rangle = \sum_{j=1}^d \sum_{k=1}^m \mathbf{A}_{jk} \mathbf{C}_{jk} = \text{tr}(\mathbf{A}^T \mathbf{C})$, where $\text{tr}(\mathbf{A})$ is the trace of a matrix $\mathbf{A}$. We use $\mathbf{A}_{*k} = (\mathbf{A}_{1k}, \ldots, \mathbf{A}_{dk})^T$ and $\mathbf{A}_{j*} = (\mathbf{A}_{j1}, \ldots, \mathbf{A}_{jm})$ to denote the $k^{\text{th}}$ column and $j^{\text{th}}$ row of $\mathbf{A}$. Let $\mathcal{S}$ be some subspace of $\mathbb{R}^{d \times m}$, we use $\mathbf{A}_\mathcal{S}$ to denote the projection of $\mathbf{A}$ onto $\mathcal{S}$, i.e., $\mathbf{A}_\mathcal{S} = \text{argmin}_{\mathbf{C} \in \mathcal{S}} ||\mathbf{C} - \mathbf{A}||_\text{F}^2$. Given a subspace $\mathcal{U} \subset \mathbb{R}^d$, we define its orthogonal complement as $\mathcal{U}_\perp = \left\{ \boldsymbol{u} \in \mathbb{R}^d \mid \boldsymbol{u}^T \boldsymbol{v} = 0, \text{ for all } \boldsymbol{v} \in \mathcal{U} \right\}$. Moreover, we define the Frobenius, spectral, and nuclear norms of $\mathbf{A}$ as $||\mathbf{A}||_\text{F} = \sqrt{\langle \mathbf{A}, \mathbf{A} \rangle}$, $||\mathbf{A}||_2 = \psi_1(\mathbf{A})$, and $||\mathbf{A}||_* = \sum_{j=1}^r \psi_j(\mathbf{A})$, where $r$ is the rank of $\mathbf{A}$, and $\psi_j(\mathbf{A})$ is the $j^{\text{th}}$ largest singular value of $\mathbf{A}$. In addition, we define the matrix block norms as $||\mathbf{A}||_{2,1} = \sum_{k=1}^m ||\mathbf{A}_{*k}||_2$, $||\mathbf{A}||_{2,\infty} = \max_{1 \leq k \leq m} ||\mathbf{A}_{*k}||_2$, $||\mathbf{A}||_{1,p} = \sum_{j=1}^d ||\mathbf{A}_{j*}||_p$, and $||\mathbf{A}||_{\infty,q} = \max_{1 \leq j \leq d} ||\mathbf{A}_{j*}||_q$, where $1 \leq p \leq \infty$ and $1 \leq q \leq \infty$. It is easy to verify that $||\mathbf{A}||_{2,1}$ and $||\mathbf{A}||_*$ are dual norms of $||\mathbf{A}||_{2,\infty}$ and $||\mathbf{A}||_2$ respectively. Let $1/\infty = 0$, then if $1/p + 1/q = 1$, $||\mathbf{A}||_{\infty,q}$ and $||\mathbf{A}||_{1,p}$ are also dual norms of each other.

## 2. Method

We solve the multivariate regression problem in (1) by the convex program

$$\widehat{\mathbf{B}} = \underset{\mathbf{B}}{\text{argmin}} \, ||\mathbf{Y} - \mathbf{XB}||_{2,1} + \lambda \mathcal{R}(\mathbf{B}), \tag{10}$$

where $\mathcal{R}(\mathbf{B})$ is a regularization function and can take the forms in (4), (5), and (6).

To understand the intuition of (10), we show that the $L_{2,1}$-loss can be viewed as a special case of the weighted square loss function. More specifically, we consider the optimization problem

$$\widehat{\mathbf{B}}^* = \underset{\mathbf{B}}{\text{argmin}} \sum_{k=1}^m \frac{1}{\sigma_k \sqrt{n}} ||\mathbf{Y}_{*k} - \mathbf{XB}_{*k}||_2^2 + \lambda \mathcal{R}(\mathbf{B}), \tag{11}$$

where $\frac{1}{\sigma_k \sqrt{n}}$ is the weight to calibrate the $k^{\text{th}}$ regression task. $\widehat{\mathbf{B}}^*$ is an "oracle" estimator (not practically calculable) since it assumes that all $\sigma_k$'s are given. Without any prior knowledge of $\sigma_k$'s, we can use the following replacement of $\sigma_k$'s,

$$\widetilde{\sigma}_k = \frac{1}{\sqrt{n}} ||\mathbf{Y}_{*k} - \mathbf{XB}_{*k}||_2, \, k = 1, \ldots, m. \tag{12}$$

We then recover (10) by replacing $\sigma_k$ in (12) by $\widetilde{\sigma}_k$. In another word, CMR calibrates different tasks by solving a regularized weighted least square problem with weights defined in (12).

## 3. Statistical Properties

For notational simplicity, we define a rescaled noise matrix $\mathbf{W} = [\mathbf{W}_{ik}] \in \mathbb{R}^{n \times m}$ with $\mathbf{W}_{ik} = \mathbf{Z}_{ik}/\sigma_k$, where $\mathbb{E}\mathbf{Z}_{ik}^2 = \sigma_k^2$ is defined in (2). Thus $\mathbf{W}$ is a random matrix with all entries having mean 0 and variance 1. We define $\mathbf{G}^0$ as the gradient of $||\mathbf{Y} - \mathbf{XB}||_{2,1}$ at $\mathbf{B} = \mathbf{B}^0$. We see that $\mathbf{G}^0$ does not depend on the unknown quantities $\sigma_k$'s since

$$\mathbf{G}_{*k}^0 = \frac{\mathbf{X}^T \mathbf{Z}_{*k}}{||\mathbf{Z}_{*k}||_2} = \frac{\mathbf{X}^T \mathbf{W}_{*k} \sigma_k}{||\mathbf{W}_{*k} \sigma_k||_2} = \frac{\mathbf{X}^T \mathbf{W}_{*k}}{||\mathbf{W}_{*k}||_2}.$$

Thus it serves as an important pivotal in our analysis. Moreover, our analysis exploits the decomposability of $\mathcal{R}(\mathbf{B})$, which is satisfied by the nuclear and $L_{1,p}$ norms (Negahban et al., 2012).

**Definition 1** *Let $\mathcal{S}$ and $\mathcal{N}$ be two subspaces of $\mathbb{R}^{d \times m}$, which are orthogonal to each other and also satisfy $\mathcal{S} \subseteq \mathcal{N}_\perp$. A regularization function $\mathcal{R}(\cdot)$ is decomposable with respect to the pair $(\mathcal{S}, \mathcal{N})$ if for any $\mathbf{A} \in \mathbb{R}^{d \times m}$, we have*

$$\mathcal{R}(\mathbf{A} + \mathbf{C}) = \mathcal{R}(\mathbf{A}) + \mathcal{R}(\mathbf{C}) \quad for \ \mathbf{A} \in \mathcal{S} \ and \ \mathbf{C} \in \mathcal{N}.$$

The decomposability of $\mathcal{R}(\mathbf{B})$ is important in analyzing the statistical properties of the estimator in (10). The next lemma shows that if we choose $\mathcal{S}$ to be some subspace of $\mathbb{R}^{d \times m}$ containing the true parameter $\mathbf{B}^0$, given a decomposable regularizer and a suitably chosen $\lambda$, the optimum to (10) lies in a restricted set.

**Lemma 2** *Let $\mathbf{B}^0 \in \mathcal{S}$ and $\widehat{\mathbf{B}}$ be an arbitrary[3] optimum to (10). We denote the estimation error as $\widehat{\mathbf{\Delta}} = \widehat{\mathbf{B}} - \mathbf{B}^0$ and the dual norm of $\mathcal{R}(\cdot)$ as $\mathcal{R}^*(\cdot)$. If $\lambda \geq c\mathcal{R}^*(\mathbf{G}^0)$ for some $c > 1$, we have*

$$\widehat{\mathbf{\Delta}} \in \mathcal{M}_c = \left\{ \mathbf{\Delta} \in \mathbb{R}^{d \times m} \mid \mathcal{R}(\mathbf{\Delta}_\mathcal{N}) \leq \frac{c+1}{c-1} \mathcal{R}(\mathbf{\Delta}_{\mathcal{N}_\perp}) \right\}. \tag{13}$$

The proof of Lemma 2 is provided in Appendix A. To prove the main result, we assume that the design matrix $\mathbf{X}$ satisfies a generalized restricted eigenvalue condition as below.

**Assumption 1** *Let $\mathbf{B}^0 \in \mathcal{S}$, then there exist positive constants $\kappa$ and $c > 1$ such that*

$$\kappa = \min_{\mathbf{\Delta} \in \mathcal{M}_c \setminus \{\mathbf{0}\}} \frac{||\mathbf{X}\mathbf{\Delta}||_{\mathrm{F}}}{\sqrt{n} ||\mathbf{\Delta}||_{\mathrm{F}}}.$$

Assumption 1 is the generalization of the restricted eigenvalue conditions for analyzing univariate sparse linear models (Negahban et al., 2012; Bickel et al., 2009). Many design matrices satisfy this assumption with high probability (Lounici et al., 2011; Negahban and Wainwright, 2011; Rohde and Tsybakov, 2011; Raskutti et al., 2010).

### 3.1 Main Result

We first present a deterministic result for a general norm-based regularization function $\mathcal{R}(\cdot)$, which satisfies the decomposability in Definition 1.

**Theorem 3** *Suppose that the design matrix $\mathbf{X}$ satisfies Assumption 1. Let $\widehat{\mathbf{B}}$ be an arbitrary optimum to (10), and $\mathbf{G}^0$ be the gradient of $||\mathbf{Y} - \mathbf{XB}||_{2,1}$ at $\mathbf{B} = \mathbf{B}^0$. We denote*

$$\Theta(\mathcal{N}_\perp, \mathcal{R}) = \max_{\mathbf{A} \in \mathbb{R}^{d \times m} \setminus \{\mathbf{0}\}} \frac{\mathcal{R}(\mathbf{A}_{\mathcal{N}_\perp})}{||\mathbf{A}_{\mathcal{N}_\perp}||_{\mathrm{F}}}.$$

*Let $\lambda$ satisfy*

$$2\lambda \Theta(\mathcal{N}_\perp, \mathcal{R}) \leq \delta(c-1)\sqrt{n}\kappa \ for \ some \ \delta < 1, \ and \ \lambda \geq c\mathcal{R}^*(\mathbf{G}^0).$$

---

3. Since (10) is not a strictly convex program, the optimum to (10) is not necessarily unique.

*Then we have*

$$\frac{1}{\sqrt{nm}}||\mathbf{X}\widehat{\mathbf{B}} - \mathbf{X}\mathbf{B}^0||_{\mathrm{F}} \leq \frac{4\lambda\Theta(\mathcal{N}_\perp, \mathcal{R})\sigma_{\max}}{\sqrt{mn}\kappa(c-1)(1-\delta)}||\mathbf{W}||_{2,\infty},$$

$$\frac{1}{\sqrt{m}}||\widehat{\mathbf{B}} - \mathbf{B}^0||_{\mathrm{F}} \leq \frac{4\lambda\Theta(\mathcal{N}_\perp, \mathcal{R})\sigma_{\max}}{\sqrt{mn}\kappa^2(c-1)(1-\delta)}||\mathbf{W}||_{2,\infty},$$

*where $\sigma_{\max} = \max\limits_{1 \leq k \leq m} \sigma_k$. Moreover, if we estimate $\sigma_k$'s by*

$$\widehat{\sigma}_k = \frac{1}{\sqrt{n}}||\mathbf{Y}_{*k} - \mathbf{X}\widehat{\mathbf{B}}_{*k}||_2 \text{ for all } k = 1, ..., m, \tag{14}$$

*then we have*

$$\frac{1}{m}\left|\sum_{k=1}^m \widehat{\sigma}_k - \sum_{k=1}^m \sigma_k\right| \leq \max\left\{1, \frac{2}{c-1}\right\}\frac{4\lambda^2\Theta^2(\mathcal{N}_\perp, \mathcal{R})\sigma_{\max}}{\sqrt{nm}n\kappa(c-1)(1-\delta)}||\mathbf{W}||_{2,\infty}.$$

The proof of Theorem 3 is provided in Appendix B. Note that Theorem 3 is a deterministic bound of the CMR estimator for a fixed $\lambda$. Since $\mathbf{W}$ is a random matrix, we need to bound $||\mathbf{W}||_{2,\infty}$ and show that $\lambda \geq c\mathcal{R}^*(\mathbf{G}^0)$ holds with high probability. For simplicity, we assume that each entry of $\mathbf{W}$ follows a Gaussian distribution as follows.

**Assumption 2** *All $\mathbf{W}_{ik}$'s are independently generated from $N(0,1)$.*

We then refine error bounds of the CMR estimator under Assumption 2 for calibrated sparse multivariate regression and calibrated low rank multivariate regression respectively .

## 3.2 Calibrated Low Rank Multivariate Regression

We assume that the rank of $\mathbf{B}^0$ is $r \ll \min\{d, m\}$, and $\mathbf{B}^0$ has a singular value decomposition $\mathbf{B}^0 = \sum_{j=1}^r \psi_j(\mathbf{B}^0)\boldsymbol{u}_j\boldsymbol{v}_j^T$ where $\psi_j(\mathbf{B}^0)$ is the $j^{\mathrm{th}}$ largest singular value with $\boldsymbol{u}_j$'s and $\boldsymbol{v}_j$'s as the corresponding left and right singular vectors. We define

$$\mathcal{U} = \mathrm{span}(\{\boldsymbol{u}_1, ..., \boldsymbol{u}_r\}) \subset \mathbb{R}^d \text{ and } \mathcal{V} = \mathrm{span}(\{\boldsymbol{v}_1, ..., \boldsymbol{v}_r\}) \subset \mathbb{R}^m.$$

We then define $\mathcal{S}$ and $\mathcal{N}$ as follows,

$$\mathcal{S} = \left\{\mathbf{C} \in \mathbb{R}^{d \times m} \mid \mathbf{C}_{*k} \in \mathcal{U}, \ \mathbf{C}_{j*} \in \mathcal{V} \text{ for all } j, k\right\}, \tag{15}$$

$$\mathcal{N} = \left\{\mathbf{C} \in \mathbb{R}^{d \times m} \mid \mathbf{C}_{*k} \in \mathcal{U}_\perp, \ \mathbf{C}_{j*} \in \mathcal{V}_\perp \text{ for all } j, k\right\}. \tag{16}$$

We can easily verify that $\mathbf{B}^0 \in \mathcal{S}$ and the nuclear norm is decomposable with respect to the pair $(\mathcal{S}, \mathcal{N})$, i.e.,

$$||\mathbf{A} + \mathbf{C}||_* = ||\mathbf{A}||_* + ||\mathbf{C}||_* \text{ for } \mathbf{A} \in \mathcal{S} \text{ and } \mathbf{C} \in \mathcal{N}.$$

The next corollary provides the concrete rates of convergence for the calibrated low rank multivariate regression estimator.

**Corollary 4** *We assume that the design matrix $\mathbf{X}$ satisfies Assumption 1 with $\mathcal{S}$ and $\mathcal{N}$ chosen as in (15) and (16), and each column of $\mathbf{X}$ is normalized so that*

$$\frac{||\mathbf{X}_{*j}||_2}{\sqrt{n}} = 1 \text{ for all } j = 1, ..., d. \tag{17}$$

We also assume that the rescaled noise matrix $\mathbf{W}$ satisfies Assumption 2. By Theorem 3, for some universal constants $c_0 \in (0, 1)$, $c_1 > 0$, and large enough $n$, we take

$$\lambda = \frac{2c\|\mathbf{X}\|_2(\sqrt{d} + \sqrt{m})}{\sqrt{n}(1 - c_0)}, \tag{18}$$

then for some $\delta < 1$, we have

$$\frac{1}{\sqrt{nm}}\|\mathbf{X}\widehat{\mathbf{B}} - \mathbf{X}\mathbf{B}^0\|_F \leq \frac{8c\sqrt{2}\|\mathbf{X}\|_2\sigma_{\max}}{\sqrt{n}\kappa(c-1)(1-\delta)}\sqrt{\frac{1+c_0}{1-c_0}}\left(\sqrt{\frac{r}{n}} + \sqrt{\frac{rd}{nm}}\right),$$

$$\frac{1}{\sqrt{m}}\|\widehat{\mathbf{B}} - \mathbf{B}^0\|_F \leq \frac{8c\sqrt{2}\|\mathbf{X}\|_2\sigma_{\max}}{\sqrt{n}\kappa^2(c-1)(1-\delta)}\sqrt{\frac{1+c_0}{1-c_0}}\left(\sqrt{\frac{r}{n}} + \sqrt{\frac{rd}{nm}}\right),$$

$$\frac{1}{m}\left|\sum_{k=1}^{m}\widehat{\sigma}_k - \sum_{k=1}^{m}\sigma_k\right| \leq \max\left\{1, \frac{2}{c-1}\right\}\frac{64c^2\|\mathbf{X}\|_2^2\sigma_{\max}}{n\kappa(c-1)(1-\delta)}\frac{\sqrt{1+c_0}}{1-c_0}\left(\frac{rd}{nm} + \frac{r}{n}\right)$$

with probability at least $1 - 2\exp(-c_1d - c_1m) - 2\exp\left(-nc_0^2/8 + \log m\right)$.

The proof of Corollary 4 is provided in Appendix C. The rate of convergence obtained in Corollary 4 matches the minimax lower bound[4] presented in Rohde and Tsybakov (2011). See more details in Theorems 5 and 6 of Rohde and Tsybakov (2011).

### 3.3 Calibrated Sparse Multivariate Regression

We now assume that the multivariate regression model in (1) is jointly sparse. More specifically, we assume that $\mathbf{B}^0$ has $s$ rows with nonzero entries and define

$$\mathcal{S} = \left\{\mathbf{C} \in \mathbb{R}^{d \times m} \mid \mathbf{C}_{j*} = \mathbf{0} \text{ for all } j \text{ such that } \mathbf{B}_{j*}^0 = \mathbf{0}\right\}, \tag{19}$$

$$\mathcal{N} = \left\{\mathbf{C} \in \mathbb{R}^{d \times m} \mid \mathbf{C}_{j*} = \mathbf{0} \text{ for all } j \text{ such that } \mathbf{B}_{j*}^0 \neq \mathbf{0}\right\}. \tag{20}$$

We can easily verify that we have $\mathbf{B}^0 \in \mathcal{S}$ and the $L_{1,p}$ norm is decomposable with respect to the pair $(\mathcal{S}, \mathcal{N})$, i.e.,

$$\|\mathbf{A} + \mathbf{C}\|_{1,p} = \|\mathbf{A}\|_{1,p} + \|\mathbf{C}\|_{1,p} \quad \text{for} \quad \mathbf{A} \in \mathcal{S} \text{ and } \mathbf{C} \in \mathcal{N}.$$

The next corollary provides the concrete rates of convergence for the calibrated sparse multivariate regression estimator.

**Corollary 5** *We assume that the design matrix $\mathbf{X}$ satisfies Assumption 1 with $\mathcal{S}$ and $\mathcal{N}$ chosen as in (19) and (20), and each column of $\mathbf{X}$ is normalized so that*

$$\frac{m^{1/2-1/p}\|\mathbf{X}_{*j}\|_2}{\sqrt{n}} = 1 \text{ for all } j = 1, ..., d. \tag{21}$$

*We also assume that the rescaled noise matrix $\mathbf{W}$ satisfies Assumption 2. By Theorem 3, for some universal constant $c_0 \in (0, 1)$ and large enough $n$, let*

$$\lambda = \frac{2c(m^{1-1/p} + \sqrt{\log d})}{\sqrt{1 - c_0}}, \tag{22}$$

---

4. In the fixed design setting for the low rank regression, $\|\mathbf{X}\|_2$ is supposed to increase as an order of $\sqrt{n}$. Thus $\|\mathbf{X}\|_2/\sqrt{n}$ in (18) should be viewed as a constant.

*then for some $\delta < 1$, we have*

$$\frac{1}{\sqrt{nm}}||\mathbf{X}\widehat{\mathbf{B}} - \mathbf{X}\mathbf{B}^0||_F \leq \frac{8c\sigma_{\max}}{\kappa(c-1)(1-\delta)}\sqrt{\frac{1+c_0}{1-c_0}}\left(\sqrt{\frac{sm^{1-2/p}}{n}} + \sqrt{\frac{s\log d}{nm}}\right),$$

$$\frac{1}{\sqrt{m}}||\widehat{\mathbf{B}} - \mathbf{B}^0||_F \leq \frac{8c\sigma_{\max}}{\kappa^2(c-1)(1-\delta)}\sqrt{\frac{1+c_0}{1-c_0}}\left(\sqrt{\frac{sm^{1-2/p}}{n}} + \sqrt{\frac{s\log d}{nm}}\right),$$

$$\frac{1}{m}\left|\sum_{k=1}^m \widehat{\sigma}_k - \sum_{k=1}^m \sigma_k\right| \leq \max\left\{1, \frac{2}{c-1}\right\}\frac{32c^2\sigma_{\max}}{\kappa(c-1)(1-\delta)}\frac{\sqrt{1+c_0}}{1-c_0}\left(\frac{sm^{1-2/p}}{n} + \frac{s\log d}{mn}\right)$$

*with probability at least $1 - 2\exp(-2\log d) - 2\exp\left(-nc_0^2/8 + \log m\right)$.*

The proof of Corollary 5 is provided in Appendix D. Note that when we choose $p = 2$, the column normalization condition (21) becomes

$$\frac{||\mathbf{X}_{*j}||_2}{\sqrt{n}} = 1 \text{ for all } j = 1, ..., d,$$

which is the same as (17). Then Corollary 5 implies that with high probability, we have

$$\frac{1}{\sqrt{m}}||\widehat{\mathbf{B}} - \mathbf{B}^0||_F \leq \frac{8c\sigma_{\max}}{\kappa^2(c-1)(1-\delta)}\sqrt{\frac{1+c_0}{1-c_0}}\left(\sqrt{\frac{s}{n}} + \sqrt{\frac{s\log d}{nm}}\right). \tag{23}$$

The rate of convergence obtained in (23) matches the minimax lower bound presented in Lounici et al. (2011). See more details in Theorem 6.1 of Lounici et al. (2011).

**Remark 6** *From Corollaries 4 and 5, we see that CMR achieves the same rates of convergence as the noncalibrated counterpart in parameter estimation. Moreover, the selected regularization parameter $\lambda$ in (18) and (22) does not involve $\sigma_k$'s. Therefore CMR makes the regularization parameter selection insensitive to $\sigma_{\max}$.*

# 4. Computational Algorithm

Though the $L_{2,1}$ norm is nonsmooth, it is nondifferentiable only when a task achieves exact zero residual, which is unlikely to happen in practice. This motivates us to apply the smoothing approach proposed by Nesterov (2005) to obtain a smooth approximation so that we can avoid directly evaluating the subgradient of the $L_{2,1}$ loss function. Thus we gain computational efficiency like other smooth loss functions.

## 4.1 Smooth Approximation

We consider the Fenchel's dual representation of the $L_{2,1}$ loss:

$$||\mathbf{Y} - \mathbf{X}\mathbf{B}||_{2,1} = \max_{||\mathbf{U}||_{2,\infty} \leq 1}\langle\mathbf{U}, \mathbf{Y} - \mathbf{X}\mathbf{B}\rangle.$$

Let $\mu > 0$ be a smoothing parameter. The smooth approximation of the $L_{2,1}$ loss can be obtained by solving the optimization problem

$$||\mathbf{Y} - \mathbf{X}\mathbf{B}||_\mu = \max_{||\mathbf{U}||_{2,\infty} \leq 1}\langle\mathbf{U}, \mathbf{Y} - \mathbf{X}\mathbf{B}\rangle - \frac{\mu}{2}||\mathbf{U}||_F^2. \tag{24}$$

Note that the equality in (24) is attained with $\mathbf{U} = \widehat{\mathbf{U}}^{\mathbf{B}}$:

$$\widehat{\mathbf{U}}^{\mathbf{B}}_{*k} = \frac{\mathbf{Y}_{*k} - \mathbf{X}\mathbf{B}_{*k}}{\max\{||\mathbf{Y}_{*k} - \mathbf{X}\mathbf{B}_{*k}||_2, \mu\}}.$$

Nesterov (2005) has shown that $||\mathbf{Y} - \mathbf{X}\mathbf{B}||_\mu$ have good computational structures: (1) It is convex and differentiable with respect to $\mathbf{B}$; (2) Its gradient takes a simple form as

$$\mathbf{G}^\mu(\mathbf{B}) = \frac{\partial ||\mathbf{Y} - \mathbf{X}\mathbf{B}||_\mu}{\partial \mathbf{B}} = \frac{\partial \left( \langle \widehat{\mathbf{U}}^{\mathbf{B}}, \mathbf{Y} - \mathbf{X}\mathbf{B} \rangle - \mu ||\widehat{\mathbf{U}}^{\mathbf{B}}||_F^2/2 \right)}{\partial \mathbf{B}} = -\mathbf{X}^T \widehat{\mathbf{U}}^{\mathbf{B}};$$

(3) Let $\gamma = ||\mathbf{X}^T\mathbf{X}||_2$, we have that $\mathbf{G}^\mu(\mathbf{B})$ is Lipschitz continuous in $\mathbf{B}$ with the Lipschitz constant $\gamma/\mu$, i.e., for any $\mathbf{B}'$, $\mathbf{B}'' \in \mathbb{R}^{d \times m}$,

$$||\mathbf{G}^\mu(\mathbf{B}') - \mathbf{G}^\mu(\mathbf{B}'')||_F \leq \frac{\gamma}{\mu}||\mathbf{B}' - \mathbf{B}''||_F.$$

Therefore we consider a smoothed replacement of the optimization problem in (10):

$$\widetilde{\mathbf{B}} = \underset{\mathbf{B}}{\mathrm{argmin}} \, ||\mathbf{Y} - \mathbf{X}\mathbf{B}||_\mu + \lambda \mathcal{R}(\mathbf{B}). \tag{25}$$

### 4.2 Smoothed Proximal Gradient Algorithm

We then present a brief derivation of the smoothed proximal gradient algorithm for solving (25). We first define three sequences of auxiliary variables $\{\mathbf{A}^{(t)}\}$, $\{\mathbf{V}^{(t)}\}$, and $\{\mathbf{H}^{(t)}\}$ with $\mathbf{A}^{(0)} = \mathbf{H}^{(0)} = \mathbf{V}^{(0)} = \mathbf{B}^{(0)}$, a sequence of weights $\{\theta_t = 2/(t+1)\}$, and a nonincreasing sequence of step sizes $\{\eta_t\}_{t=0}^\infty$.

At the $t^{\text{th}}$ iteration, we take $\mathbf{V}^{(t)} = (1 - \theta_t)\mathbf{B}^{(t-1)} + \theta_t\mathbf{A}^{(t-1)}$. Let $\widetilde{\mathbf{H}}^{(t)} = \mathbf{V}^{(t)} - \eta_t\mathbf{G}^\mu(\mathbf{V}^{(t)})$. When $\mathcal{R}(\mathbf{H}) = ||\mathbf{H}||_*$, we take

$$\mathbf{H}^{(t)} = \sum_{j=1}^{\min\{d,m\}} \max\left\{\psi_j(\widetilde{\mathbf{H}}^{(t)}) - \eta_t\lambda, 0\right\} \boldsymbol{u}_j \boldsymbol{v}_j^T,$$

where $\boldsymbol{u}_j$ and $\boldsymbol{v}_j$ are the left and right singular vectors of $\widetilde{\mathbf{H}}^{(t)}$ corresponding to the $j^{\text{th}}$ largest singular value $\psi_j(\widetilde{\mathbf{H}}^{(t)})$. When $\mathcal{R}(\mathbf{H}) = ||\mathbf{H}||_{1,2}$, we take

$$\mathbf{H}^{(t)}_{j*} = \widetilde{\mathbf{H}}_{j*} \cdot \max\left\{1 - \eta_t\lambda/||\widetilde{\mathbf{H}}_{j*}||_2, 0\right\}.$$

See more details about other choices of $p$ in the $L_{1,p}$ norm in Liu et al. (2009a); Liu and Ye (2010). To ensure that the objective function value is nonincreasing, we choose

$$\mathbf{B}^{(t)} = \underset{\mathbf{B} \in \{\mathbf{H}^{(t)}, \, \mathbf{B}^{(t-1)}\}}{\mathrm{argmin}} \, ||\mathbf{Y} - \mathbf{X}\mathbf{B}||_\mu + \lambda \mathcal{R}(\mathbf{B}).$$

For simplicity, we can set $\{\eta_t\}$ as a constant sequence, e.g., $\eta_t = \mu/\gamma$ for $t = 1, 2, \dots$. In practice, we cam use the backtracking line search to adjust $\eta_t$ and boost the performance. At last, we take $\mathbf{A}^{(t)} = \mathbf{B}^{(t-1)} + \frac{1}{\theta_t}(\mathbf{H}^{(t)} - \mathbf{B}^{(t-1)})$. Given a stopping precision $\varepsilon$, the algorithm stops when $\max\left\{||\mathbf{B}^{(t)} - \mathbf{B}^{(t-1)}||_F, \, ||\mathbf{H}^{(t)} - \mathbf{H}^{(t-1)}||_F\right\} \leq \varepsilon$.

**Remark 7** *The smoothed proximal gradient algorithm has a worst-case iteration complexity of $\mathcal{O}(1/\epsilon)$, where $\epsilon$ is a pre-specified accuracy of the objective value*[5]. *See more details in Nesterov (2005); Beck and Teboulle (2009a).*

---

5. During this paper was under review, a dual proximal gradient algorithm was proposed for solving (10). See more details in Gong et al. (2014).

## 5. Numerical Experiments

To compare the finite-sample performance between the calibrated multivariate regression (CMR) and ordinary multivariate regression (OMR), we conduct numerical experiments on both simulated and real data sets.

### 5.1 Simulated Data

We generate training data sets of 400 samples for the low rank setting and 200 samples for joint sparsity setting. In details, for the low rank setting, we use the following data generation scheme:

(1) Generate each row of the design matrix $\mathbf{X}_{i*}$, $i = 1, ..., 400$, independently from a 200-dimensional normal distribution $N(\mathbf{0}, \boldsymbol{\Sigma})$ where $\boldsymbol{\Sigma}_{jj} = 1$ and $\boldsymbol{\Sigma}_{j\ell} = 0.5$ for all $\ell \neq j$.

(2) Generate the regression coefficient matrix $\mathbf{B}^0 = \mathbf{L}\mathbf{R}^T$, where $\mathbf{L} \in \mathbb{R}^{200 \times 3}$, $\mathbf{R} \in \mathbb{R}^{3 \times 101}$, and all entries of $\mathbf{L}$ and $\mathbf{R}$ are independently generated from $N(0, 0.05)$.

(3) Generate the random noise matrix $\mathbf{Z} = \mathbf{W}\mathbf{D}$ where $\mathbf{W} \in \mathbb{R}^{400 \times 101}$ with all entries of $\mathbf{W}$ independently generated from $N(0, 1)$ and $\mathbf{D}$ is either of the following matrices

$$\mathbf{D} = \sigma_{\max} \cdot \operatorname{diag}\left(2^{0/100}, 2^{-3/100}, \cdots, 2^{-297/100}, 2^{-300/100}\right) \in \mathbb{R}^{101 \times 101},, \tag{26}$$

$$\mathbf{D} = \sigma_{\max} \cdot \operatorname{diag}\left(1, 1, \cdots, 1, 1\right) \in \mathbb{R}^{101 \times 101}. \tag{27}$$

For the joint sparsity setting, we use the following data generation scheme:

(1) Generate each row of the design matrix $\mathbf{X}_{i*}$, $i = 1, ..., 200$, independently from a 800-dimensional normal distribution $N(\mathbf{0}, \boldsymbol{\Sigma})$ where $\boldsymbol{\Sigma}_{jj} = 1$ and $\boldsymbol{\Sigma}_{j\ell} = 0.5$ for all $\ell \neq j$.

(2) Let $k = 1, \ldots, 13$, set the regression coefficient matrix $\mathbf{B}^0 \in \mathbb{R}^{800 \times 13}$ as $\mathbf{B}^0_{1k} = 3$, $\mathbf{B}^0_{2k} = 2$, $\mathbf{B}^0_{4k} = 1.5$, and $\mathbf{B}^0_{jk} = 0$ for all $j \neq 1, 2, 4$.

(3) Generate the random noise matrix $\mathbf{Z} = \mathbf{W}\mathbf{D}$, where $\mathbf{W} \in \mathbb{R}^{200 \times 13}$ with all entries of $\mathbf{W}$ independently generated from $N(0, 1)$ and $\mathbf{D}$ is is either of the following matrices

$$\mathbf{D} = \sigma_{\max} \cdot \operatorname{diag}\left(2^{0/4}, 2^{-1/4}, \cdots, 2^{-11/4}, 2^{-12/4}\right) \in \mathbb{R}^{13 \times 13}, \tag{28}$$

$$\mathbf{D} = \sigma_{\max} \cdot \operatorname{diag}\left(1, 1, \cdots, 1, 1\right) \in \mathbb{R}^{13 \times 13}. \tag{29}$$

In addition, we generate validation sets (400 samples for the low rank setting and 200 samples for the joint sparsity setting) for the regularization parameter selection, and testing sets (10,000 samples for both settings) to evaluate the prediction accuracy.

**Remark 8** *The scale matrices in (26) and (28) consider the scenario, where the regression tasks have different variances. The scale matrices in (27) and (29) consider the scenario, where all regression tasks have the equal variance.*

In numerical experiments, we set $\sigma_{\max} = 1$, 2, and 4 to illustrate the tuning insensitivity of CMR. The regularization parameter $\lambda$ of both CMR and OMR is chosen over a grid

$$\boldsymbol{\Lambda} = \left\{2^{40/4}\lambda_0, 2^{39/4}\lambda_0, \cdots, 2^{-17/4}\lambda_0, 2^{-18/4}\lambda_0\right\}.$$

We choose

$$\lambda_0 = \frac{\|\mathbf{X}\|_2}{n}(\sqrt{d} + \sqrt{m}) \quad \text{and} \quad \lambda_0 = \sqrt{\log d} + \sqrt{m}$$

for the low rank and joint sparsity settings. The optimal regularization parameter $\widehat{\lambda}$ is determined by the prediction error as

$$\widehat{\lambda} = \underset{\lambda \in \Lambda}{\operatorname{argmin}} \, ||\widetilde{\mathbf{Y}} - \widetilde{\mathbf{X}}\widehat{\mathbf{B}}^{\lambda}||_{\mathrm{F}}^2,$$

where $\widehat{\mathbf{B}}^{\lambda}$ denotes the obtained estimate using the regularization parameter $\lambda$, and $\widetilde{\mathbf{X}}$ and $\widetilde{\mathbf{Y}}$ denote the design and response matrices of the validation set.

Since the noise level $\sigma_k$'s may vary across different regression tasks, we adopt the following three criteria to evaluate the empirical performance:

$$\text{P.E.} = \frac{1}{10000}||\overline{\mathbf{Y}} - \overline{\mathbf{X}}\widehat{\mathbf{B}}||_{\mathrm{F}}^2, \quad \text{A.P.E.} = \frac{1}{10000m}||(\overline{\mathbf{Y}} - \overline{\mathbf{X}}\widehat{\mathbf{B}})\mathbf{D}^{-1}||_{\mathrm{F}}^2, \quad \text{E.E.} = \frac{1}{m}||\widehat{\mathbf{B}} - \mathbf{B}^0||_{\mathrm{F}}^2,$$

where $\overline{\mathbf{X}}$ and $\overline{\mathbf{Y}}$ denote the design and response matrices of the testing set.

All simulations are implemented by MATLAB using a PC with Intel Core i5 3.3GHz CPU and 16GB memory. We set $p = 2$ for the joint sparsity setting, but it is straightforward to extend to arbitrary $p > 2$. OMR is solved by the monotone fast proximal gradient algorithm, where we set the stopping precision $\varepsilon = 10^{-4}$. CMR is solved by the proposed smoothed proximal gradient algorithm, where we set the stopping precision $\varepsilon = 10^{-4}$, and the smoothing parameter $\mu = 10^{-4}$.

We compare the statistical performance between CMR and OMR. Tables 1-4 summarize the results averaged over 200 simulations for both settings. In addition, since we know the true values of $\sigma_k$'s, we also present the results of the oracle estimator $\widehat{\mathbf{B}}^*$ defined in (11). The oracle estimator is only for comparison purpose, and it is not a practical estimator.

Tables 1 and 3 present the empirical results when we adopt the scale matrix $\mathbf{D}$ defined in (26) and (28) to generate the random noise. Though our theoretical analysis in §3 only shows CMR attains the same rates of convergence as OMR, our empirical results show that CMR universally outperforms OMR, and achieves almost the same performance as the oracle estimator. These results corroborate the effectiveness of the calibration for each task.

| $\sigma_{\max}$ | Method | P.E. | A.P.E. | E.E. |
|---|---|---|---|---|
| | Oracle | 48.394(0.7421) | 1.1659(0.0241) | 0.1106(0.0245) |
| 1 | CMR | 48.411(0.7431) | 1.1668(0.0214) | 0.1109(0.0133) |
| | OMR | 53.337(0.7063) | 1.2880(0.0231) | 0.2077(0.0137) |
| | Oracle | 183.38(0.9786) | 1.0917(0.0068) | 0.2425(0.0187) |
| 2 | CMR | 183.40(1.2212) | 1.0924(0.0063) | 0.2430(0.0238) |
| | OMR | 194.66(1.4109) | 1.1641(0.0112) | 0.4637(0.0277) |
| | Oracle | 713.13(3.3923) | 1.0554(0.0062) | 0.5696(0.0669) |
| 4 | CMR | 713.24(2.7685) | 1.0565(0.0047) | 0.5737(0.0533) |
| | OMR | 728.55(2.6500) | 1.0793(0.0051) | 0.8722(0.0526) |

Table 1: Quantitative comparison of the statistical performance between CMR and OMR for the low rank setting with $\mathbf{D}$ defined in (26). The results are averaged over 200 simulations with the standard errors in parentheses. CMR universally outperforms OMR, and achieves almost the same performance as the oracle estimator.

Tables 2 and 4 present the empirical results when we adopt the scale matrix $\mathbf{D}$ defined in (27) and (29) with all $\sigma_k$'s being equal. We can see that CMR attains similar performance to OMR. This indicates that CMR is a safe replacement of OMR for multivariate regressions.

| $\sigma_{max}$ | Method | P.E. | A.P.E. | E.E. |
|---|---|---|---|---|
| 1 | CMR | 112.13(1.0051) | 1.1103(0.0097) | 0.2128(0.0190) |
|   | OMR | 113.69(1.0163) | 1.1951(0.0100) | 0.2217(0.0193) |
| 2 | CMR | 428.42(1.8061) | 1.0605(0.0043) | 0.4576(0.0344) |
|   | OMR | 430.73(1.9636) | 1.0758(0.0053) | 0.4752(0.0413) |
| 4 | CMR | 1669.2(5.0401) | 1.0335(0.0028) | 0.9621(0.0991) |
|   | OMR | 1673.5(5.7879) | 1.0378(0.0035) | 1.0353(0.1104) |

Table 2: Quantitative comparison of the statistical performance between CMR and OMR for the low rank setting with $\mathbf{D}$ defined in (27). The results are averaged over 200 simulations with the standard errors in parentheses. CMR and OMR achieve similar statistical performance.

| $\sigma_{max}$ | Method | P.E. | A.P.E. | E.E. |
|---|---|---|---|---|
| 1 | Oracle | 5.8759(0.0834) | 1.0454(0.0149) | 0.0245(0.0086) |
|   | CMR | 5.8761(0.0669) | 1.0459(0.0122) | 0.0249(0.0078) |
|   | OMR | 5.9012(0.0701) | 1.0581(0.0162) | 0.0290(0.0091) |
| 2 | Oracle | 23.464(0.3237) | 1.0441(0.0148) | 0.0926(0.0342) |
|   | CMR | 23.465(0.2600) | 1.0446(0.0131) | 0.0928(0.0268) |
|   | OMR | 23.580(0.2832) | 1.0573(0.0170) | 0.1115(0.0365) |
| 4 | Oracle | 93.532(0.8843) | 1.0418(0.0962) | 0.3342(0.1255) |
|   | CMR | 93.542(0.9788) | 1.0421(0.0113) | 0.3346(0.1002) |
|   | OMR | 94.094(1.0978) | 1.0550(0.0166) | 0.4125(0.1417) |

Table 3: Quantitative comparison of the statistical performance between CMR and OMR for the joint sparsity setting with $\mathbf{D}$ defined in (28). The results are averaged over 200 simulations with the standard errors in parentheses. CMR universally outperforms OMR, and achieves almost the same performance as the oracle estimator.

| $\sigma_{max}$ | Method | P.E. | A.P.E. | E.E. |
|---|---|---|---|---|
| 1 | CMR | 13.565(0.1411) | 1.0435(0.0156) | 0.0599(0.0199) |
|   | OMR | 13.697(0.1554) | 1.0486(0.0142) | 0.0607(0.0128) |
| 2 | CMR | 54.171(0.5791) | 1.0418(0.0101) | 0.2252(0.0644) |
|   | OMR | 54.221(0.6173) | 1.0427(0.0118) | 0.2359(0.0821) |
| 4 | CMR | 215.98(1.994) | 1.0384(0.0099) | 0.80821(0.2417) |
|   | OMR | 216.19(2.391) | 1.0394(0.0114) | 0.81957(0.3180) |

Table 4: Quantitative comparison of the statistical performance between CMR and OMR for the joint sparsity setting with $\mathbf{D}$ defined in (29). The results are averaged over 200 simulations with the standard errors in parentheses. CMR and OMR achieve similar statistical performance.

In addition, we also examine the optimal regularization parameters for CMR and OMR over all replicates. We visualize the distribution of all 200 selected $\widehat{\lambda}$'s using the kernel density estimator. In particular, we adopt the Gaussian kernel, and select the kernel bandwidth based on the 10-fold cross validation. Figure 1 illustrates the estimated density functions. The horizonal axis corresponds to the rescaled regularization parameter as follows:

$$\text{Low Rank} \quad : \quad \log\left(\frac{\widehat{\lambda}}{(\sqrt{d} + \sqrt{m})\|\mathbf{X}\|_2/n}\right),$$

$$\text{Joint Sparsity} \quad : \quad \log\left(\frac{\widehat{\lambda}}{\sqrt{\log d} + \sqrt{m}}\right).$$

We see that the optimal regularization parameters of OMR significantly vary with different $\sigma_{\max}$. In contrast, the optimal regularization parameters of CMR are more concentrated. This is consistent with our claimed tuning insensitivity.

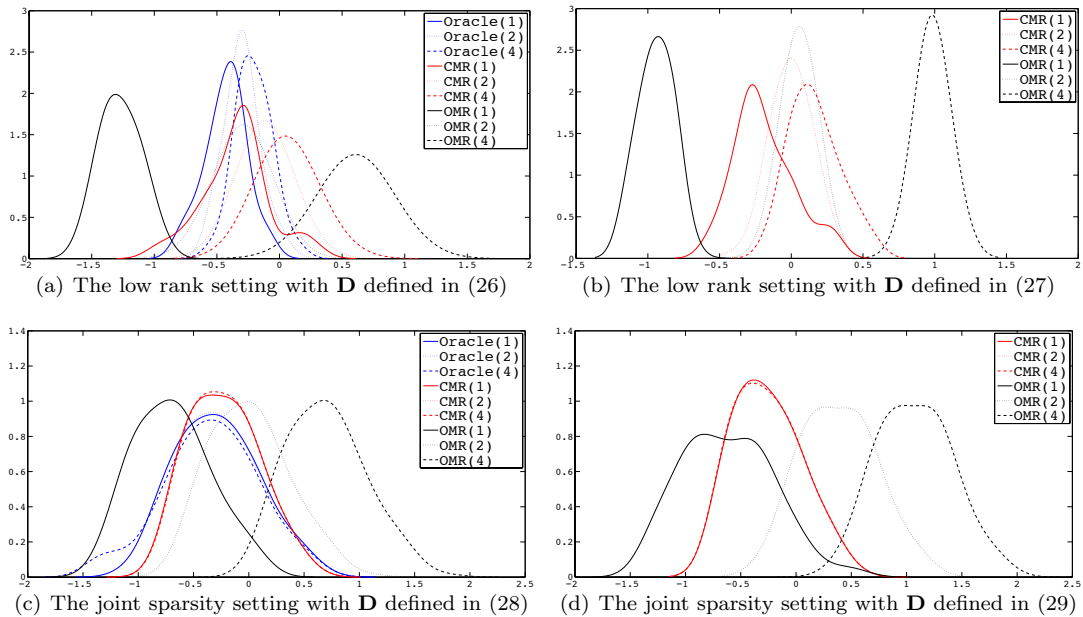

(a) The low rank setting with $\mathbf{D}$ defined in (26)

(b) The low rank setting with $\mathbf{D}$ defined in (27)

(c) The joint sparsity setting with $\mathbf{D}$ defined in (28)

(d) The joint sparsity setting with $\mathbf{D}$ defined in (29)

Figure 1: The distributions of the selected regularization parameters using the kernel density estimator. The numbers in the parentheses are $\sigma_{\max}$'s. The optimal regularization parameters of OMR are more spread with different $\sigma_{\max}$ than those of CMR and the oracle estimator.

## 5.2 Real Data

We apply CMR on a brain activity prediction problem which aims to build a parsimonious model to predict a person's neural activity when seeing a stimulus word. As is illustrated in Figure 2, for a given stimulus word, we first encode it into an intermediate semantic feature vector using some corpus statistics. We then model the brain's neural activity pattern using CMR. Creating such a predictive model not only enables us to explore new analytical tools for the fMRI data, but also helps us to gain deeper understanding on how human brain represents knowledge (Mitchell et al., 2008). As will be shown in the section, prediction based on the features selected by CMR significantly outperforms that based on the features selected by OMR, and is even better than that based on the handcrafted features selected by human experts.

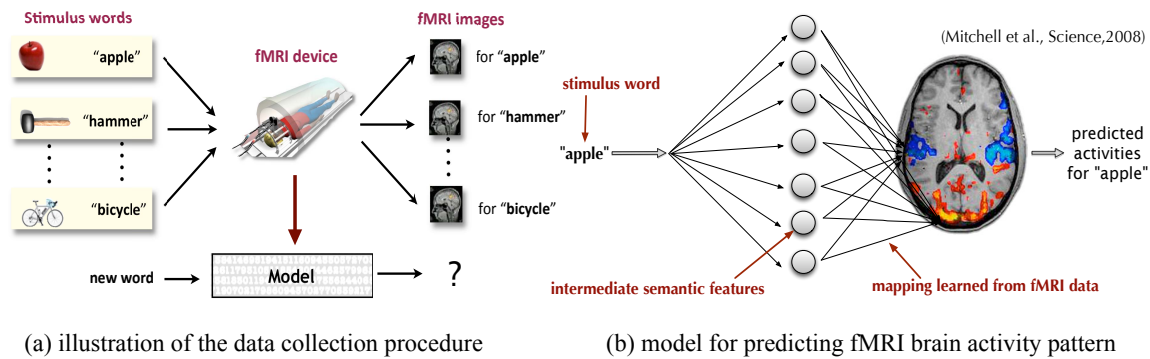

(a) illustration of the data collection procedure  (b) model for predicting fMRI brain activity pattern

Figure 2: An illustration of the fMRI brain activity prediction problem (Mitchell et al., 2008). (a) To collect the data, a human participant sees a sequence of English words and their images. The corresponding fMRI images are recorded to represent the brain activity patterns; (b) To build a predictive model, each stimulus word is encoded into intermediate semantic features (e.g. the co-occurrence statistics of this stimulus word in a large text corpus). These intermediate features can then be used to predict the brain activity pattern.

### 5.2.1 DATA

The data are obtained from Mitchell et al. (2008) and contain a fMRI image data set and a text data set. The fMRI data are collected from an experiment with 9 participants.60 nouns are selected as stimulus words from 12 different categories (See Table 5). When a participant sees a stimulus word, the fMRI device records an image[6]. Each image contains 20,601 voxels that represent the neural activities of the participant's brain. Therefore the total number of images is $9 \times 60 = 540$. Since many of the 20,601 voxels are noisy, Mitchell et al. (2008) exploit a "stability score" approach to extract 500 most stable voxels. See more details in Mitchell et al. (2008).

The text data set is collected from the Google Trillion Word corpus[7]. It contains the co-occurrence frequencies of the 60 stimulus words with 5,000 most frequent English words in the corpus with 100 stop words removed. In Mitchell et al. (2008), 25 sensory-action verbs (See Table 6) are handcrafted by human experts based on the domain knowledge of cognitive neuroscience. These 25 words are closely related to the 60 stimulus words in their semantics meanings. For example, "eat" is related to vegetables such as "lettuce" or "tomato", and "wear" is related to clothing such as "shirt" and "dress".

When building multivariate linear models, Mitchell et al. (2008) use the co-occurrence frequencies of each stimulus word with 25 sensory verbs as covariates and use the corresponding fMRI image as response. They estimate a 25-dimensional multivariate linear model by the ridge regression. They show that the obtained predictive model significantly outperforms random guess. Thus, they treat these 25 words as a semantic basis.

In our experiment below, we apply CMR to automatically select a semantic basis from all 5,000 most frequent English words. Compared with the protocol used in Mitchell et al. (2008), our approach is completely data-driven and outperforms the handcraft method in the brain activity prediction accuracy for 5 out of 9 participants.

---

6. Each image is actually the average of 6 consecutive recordings of each word.

7. http://googleresearch.blogspot.com/2006/08/all-our-n-gram-are-belong-to-you.html

| Category | Exemplar 1 | Exemplar 2 | Exemplar 3 | Exemplar 4 | Exemplar 5 |
|---|---|---|---|---|---|
| animals | bear | cat | cow | dog | horse |
| body parts | arm | eye | foot | hand | leg |
| buildings | apartment | barn | church | house | igloo |
| building parts | arch | chimney | closet | door | window |
| clothing | coat | dress | pants | shirt | skirt |
| furniture | bed | chair | desk | dresser | table |
| insects | ant | bee | beetle | butterfly | fly |
| kitchen utensils | bottle | cup | glass | knife | spoon |
| man made objects | bell | key | refrigerator | telephone | watch |
| tools | chisel | hammer | pliers | saw | screwdriver |
| vegetables | carrot | celery | corn | lettuce | tomato |
| vehicles | airplane | bicycle | car | train | truck |

Table 5: The 60 stimulus words used in Mitchell et al. (2008) from 12 categories (5 per category).

| | | | | |
|---|---|---|---|---|
| See | Eat | Run | Say | Enter |
| Hear | Touch | Push | Fear | Drive |
| Listen | Rub | Fill | Open | Wear |
| Taste | Approach | Move | Lift | Break |
| Smell | Manipulate | Ride | Near | Clean |

Table 6: The 25 verbs used in Mitchell et al. (2008). They are handcrafted based on the domain knowledge of cognitive science, and are independent on the data set.

### 5.2.2 Experimental Protocol in Mitchell et al. (2008)

The evaluation procedure of Mitchell et al. (2008) is based on the leave-two-out cross validation over all $\binom{60}{2} = 1,770$ possible partitions. In each partition, we select 58 stimulus words out of 60 as the training set. Recall that each stimulus word is represented by 5,000 features and each feature is the co-occurrence frequency of a potential basis word with the stimulus word, we obtain a $58 \times 5,000$ design matrix. Similarly, we can format the fMRI images corresponding to the 58 training stimulus words into a $58 \times 500$ response matrix. In the training stage, we apply CMR and OMR to select 25 basis words by adjusting the regularization parameters. We then use the remaining two stimulus words as a validation set and apply the estimated models to predict the neural activity of these two stimulus words. We evaluate the prediction performance based on the combined cosine similarity measure defined as follow.

**Definition 9 (Combined Similarity Measure, Mitchell et al. (2008))** *Let $\boldsymbol{u} \in \mathbb{R}^m$ and $\boldsymbol{v} \in \mathbb{R}^m$ denote the observed fMRI images of two stimulus words in the validation set, and $\widehat{\boldsymbol{u}} \in \mathbb{R}^m$ and $\widehat{\boldsymbol{v}} \in \mathbb{R}^m$ denote the corresponding predicted fMRI images. We say that the predicted images $\widehat{\boldsymbol{u}}$ and $\widehat{\boldsymbol{v}}$ correctly label two validation stimulus words, if*

$$\cos(\boldsymbol{u}, \widehat{\boldsymbol{u}}) + \cos(\boldsymbol{v}, \widehat{\boldsymbol{v}}) > \cos(\boldsymbol{u}, \widehat{\boldsymbol{v}}) + \cos(\boldsymbol{v}, \widehat{\boldsymbol{u}}), \tag{30}$$

*where $\cos(\boldsymbol{u}, \boldsymbol{v}) = (\boldsymbol{u}^T \boldsymbol{v})/(||\boldsymbol{u}||_2 ||\boldsymbol{v}||_2)$.*

We then summarize the overall prediction accuracy for each participant by the percentage of the correct labelings over all 1,770 partitions. Table 7 presents the prediction accuracies for the 9 participants. We see that CMR universally outperforms OMR across all 9 participants by 4.42% on average. Note that the statistically significant accuracy at 5% level is 0.61, CMR achieves statistically significant advantages for 8 out of 9 participants.

| Method | P. 1 | P. 2 | P. 3 | P. 4 | P. 5 | P. 6 | P. 7 | P. 8 | P. 9 |
|--------|-------|-------|-------|-------|-------|-------|-------|-------|-------|
| CMR | 0.783 | 0.724 | 0.748 | 0.528 | 0.772 | 0.713 | 0.728 | 0.739 | 0.763 |
| OMR | 0.749 | 0.685 | 0.732 | 0.485 | 0.724 | 0.661 | 0.688 | 0.682 | 0.693 |

Table 7: Prediction accuracies evaluated using the experimental protocol in Mitchell et al. (2008). CMR universally outperforms OMR across all participants.

### 5.2.3 An Improved Experimental Protocol

There are two drawbacks of the previous protocol: (1) The selected basis words vary a lot across different partitions of the cross validation and participants. Such high variability makes the obtained results difficult to interpret; (2) The automatic semantic basis selection method of CMR and OMR is sensitive to data outliers, which are common in fMRI studies. In this section, we improve this protocol to address these two problems in a more data-driven manner.

Our main idea is to simultaneously exploit the training data of multiple participants and use the stability criterion to select more stable semantic basis words (Meinshausen and Bühlmann, 2010). In detail, for each participant to be evaluated, we choose three other representatives out of the remaining eight according to who achieve the best three leave-two-out cross validation prediction accuracies in Table 7. Taking Participant 2 and CMR as an example, the three selected representatives are Participants 1, 3, and 9 with the three highest accuracies of 0.783, 0.772, and 0.763. In this way, we could eliminate the effects of possible data outliers. We then combine the fMRI images of three representatives and formulate a multivariate regression problem with 1,500 dimensional response. We conduct the leave-two-out cross validation as in the previous protocol using the combined data set, and count the frequency of each potential basis word that appears in all 1,770 partitions. We then choose the 25 most frequent words as the semantic basis. Finally, we apply the same procedure as in the previous protocol on the current candidate participant and evaluate the prediction accuracy using the combined cosine score.

Table 8 summarizes the prediction performance based on this improved protocol. We also report the results obtained by the 25 handcrafted basis. Compared with the results in Table 7, we see that the performance of CMR is greatly improved. For Participants 1, 2, 3, 5, and 8, the prediction performance of CMR significantly outperforms the handcraft method. Moreover, since the candidate participant is not involved in the semantic basis word selection, our results imply that the selected semantic basis have good generalization capability across participants.

| Method | P. 1 | P. 2 | P. 3 | P. 4 | P. 5 | P. 6 | P. 7 | P. 8 | P. 9 |
|--------|-------|-------|-------|-------|-------|-------|-------|-------|-------|
| CMR | 0.840 | 0.794 | 0.861 | 0.651 | 0.823 | 0.722 | 0.738 | 0.720 | 0.780 |
| OMR | 0.803 | 0.789 | 0.801 | 0.602 | 0.766 | 0.623 | 0.726 | 0.749 | 0.765 |
| Handcraft | 0.822 | 0.776 | 0.773 | 0.727 | 0.782 | 0.865 | 0.734 | 0.685 | 0.819 |

Table 8: Prediction accuracies evaluated used a more heuristic protocol. CMR significantly outperforms the handcrafted basis words for 5 out of 9 participants.

Table 9 lists 35 basis words obtained by CMR using the improved protocol. The words in the bold font are common ones shared by all 9 participants. We see that our list contains nouns, adjectives, and verbs. These words are closely related to the 60 stimulus words. For example, lodge, hotel, and floor are closely related to "building" and "building parts"; green and fruit clearly refer to words in "vegetable"; built and using are related to "tools" and "man made objects".

| | | | | | |
|---|---|---|---|---|---|
| **av** | balls | booking | **built** | cartoon | cream |
| cut | **country** | **discounts** | floor | fruit | green |
| **hold** | holidays | **hotel** | interior | kill | liquid |
| located | lodge | **log** | measure | **mesh** | **near** |
| **offers** | put | **reg** | room | sale | **separate** |
| shipping | **soft** | usd | **using** | went | |

Table 9: The 35 basis words selected by CMR using the improved protocol. The words in the bold font are shared by predictive models for all 9 participants.

## 6. Discussion and Conclusion

Two other related methods are the square-root low rank multivariate regression (Klopp, 2011) and the square-root sparse multivariate regression (Bunea et al., 2013). They solve the convex program

$$\widehat{\mathbf{B}} = \underset{\mathbf{B}}{\operatorname{argmin}} \|\mathbf{Y} - \mathbf{X}\mathbf{B}\|_{\mathrm{F}} + \lambda \mathcal{R}(\mathbf{B}). \tag{31}$$

The Frobenius loss in (31) makes the regularization parameter selection independent of $\sigma_{\max}$, but it does not calibrate different regression tasks. We can rewrite (31) as

$$(\widehat{\mathbf{B}}, \widehat{\sigma}) = \underset{\mathbf{B}, \sigma}{\operatorname{argmin}} \ \frac{1}{\sqrt{nm}\sigma} \|\mathbf{Y} - \mathbf{X}\mathbf{B}\|_{\mathrm{F}}^2 + \lambda \mathcal{R}(\mathbf{B}) \ \text{ subject to } \sigma = \frac{1}{\sqrt{nm}} \|\mathbf{Y} - \mathbf{X}\mathbf{B}\|_{\mathrm{F}}. \tag{32}$$

Since $\sigma$ in (32) is not specific to any individual task, it cannot calibrate the regularization. Thus it is fundamentally different from CMR.

The calibration technique proposed in this paper is quite general, and can be extended to more sophisticated scenarios, e.g. the regularization function is weakly decomposable or geometrically decomposable (Geer, 2014; Lee et al., 2013), or the regression coefficient matrix can be decomposed into multiple structured matrices (Agarwal et al., 2012; Chen et al., 2011; Gong et al., 2012; Jalali et al., 2010; Obozinski et al., 2010). Accordingly, the extensions of our proposed theory are also straightforward. We only need to replace their squared Frobenius loss-based analysis with the $L_{2,1}$ loss based analysis in this paper.

## Appendix A. Proof of Lemma 2

Note that the following two relations are frequently used in our analysis,

$$\mathbf{Y} - \mathbf{X}\mathbf{B}^0 = \mathbf{X}\mathbf{B}^0 + \mathbf{Z} - \mathbf{X}\mathbf{B}^0 = \mathbf{Z} \quad \text{and} \quad \mathbf{Y} - \mathbf{X}\widehat{\mathbf{B}} = \mathbf{X}\mathbf{B}^0 + \mathbf{Z} - \mathbf{X}\widehat{\mathbf{B}} = \mathbf{Z} - \mathbf{X}\widehat{\boldsymbol{\Delta}}.$$

**Proof** Since $\mathbf{B}^0 \in \mathcal{S}$, we have $\mathbf{B}^0_{\mathcal{S}_\perp} = \mathbf{0}$. Then we have

$$\mathcal{R}(\widehat{\mathbf{B}}) = \mathcal{R}(\mathbf{B}^0 + \widehat{\boldsymbol{\Delta}}) = \mathcal{R}(\mathbf{B}^0_{\mathcal{S}} + \widehat{\boldsymbol{\Delta}}_{\mathcal{N}_\perp} + \widehat{\boldsymbol{\Delta}}_{\mathcal{N}}) \geq \mathcal{R}(\mathbf{B}^0_{\mathcal{S}} + \widehat{\boldsymbol{\Delta}}_{\mathcal{N}}) - \mathcal{R}(\widehat{\boldsymbol{\Delta}}_{\mathcal{N}_\perp}). \tag{33}$$

Since $\mathcal{R}(\cdot)$ is decomposable with respect to $(\mathcal{S}, \mathcal{N})$, (33) further implies

$$\mathcal{R}(\widehat{\mathbf{B}}) \geq \mathcal{R}(\mathbf{B}_{\mathcal{S}}^0) + \mathcal{R}(\widehat{\boldsymbol{\Delta}}_{\mathcal{N}}) - \mathcal{R}(\widehat{\boldsymbol{\Delta}}_{\mathcal{N}_\perp}). \tag{34}$$

Since $\mathbf{B}^0 \in \mathcal{S}$, we have $\mathcal{R}(\mathbf{B}^0) = \mathcal{R}(\mathbf{B}_{\mathcal{S}}^0)$. Then by rearranging (34), we obtain

$$\mathcal{R}(\mathbf{B}^0) - \mathcal{R}(\widehat{\mathbf{B}}) \leq \mathcal{R}(\widehat{\boldsymbol{\Delta}}_{\mathcal{N}_\perp}) - \mathcal{R}(\widehat{\boldsymbol{\Delta}}_{\mathcal{N}}). \tag{35}$$

Since $\widehat{\mathbf{B}}$ is the optimum to (10), by (34), we further have

$$||\mathbf{X}\widehat{\boldsymbol{\Delta}} - \mathbf{Z}||_{2,1} - ||\mathbf{Z}||_{2,1} \leq \lambda(\mathcal{R}(\mathbf{B}^0) - \mathcal{R}(\mathbf{B}^0 + \widehat{\boldsymbol{\Delta}})) \leq \lambda(\mathcal{R}(\widehat{\boldsymbol{\Delta}}_{\mathcal{N}_\perp}) - \mathcal{R}(\widehat{\boldsymbol{\Delta}}_{\mathcal{N}})). \tag{36}$$

Due to the convexity of $|| \cdot ||_{2,1}$, we know

$$||\mathbf{X}\widehat{\boldsymbol{\Delta}} - \mathbf{Z}||_{2,1} - ||\mathbf{Z}||_{2,1} \geq \langle \mathbf{G}^0, \widehat{\boldsymbol{\Delta}} \rangle \geq -|\langle \mathbf{G}^0, \widehat{\boldsymbol{\Delta}} \rangle|. \tag{37}$$

By the Cauchy-Schwarz inequality, we obtain

$$|\langle \mathbf{G}^0, \widehat{\boldsymbol{\Delta}} \rangle| \leq \mathcal{R}^*(\mathbf{G}^0)\mathcal{R}(\widehat{\boldsymbol{\Delta}}) \leq \frac{\lambda}{c}(\mathcal{R}(\widehat{\boldsymbol{\Delta}}_{\mathcal{N}_\perp}) + \mathcal{R}(\widehat{\boldsymbol{\Delta}}_{\mathcal{N}})), \tag{38}$$

where the last inequality comes from the assumption $\lambda \geq c\mathcal{R}^*(\mathbf{G}^0)$ and the triangle inequality $\mathcal{R}(\widehat{\boldsymbol{\Delta}}) \leq \mathcal{R}(\widehat{\boldsymbol{\Delta}}_{\mathcal{N}_\perp}) + \mathcal{R}(\widehat{\boldsymbol{\Delta}}_{\mathcal{N}})$. By combining (36), (37), and (38), we obtain

$$-\frac{\lambda}{c}(\mathcal{R}(\widehat{\boldsymbol{\Delta}}_{\mathcal{N}_\perp}) + \mathcal{R}(\widehat{\boldsymbol{\Delta}}_{\mathcal{N}})) \leq \lambda(\mathcal{R}(\widehat{\boldsymbol{\Delta}}_{\mathcal{N}_\perp}) - \mathcal{R}(\widehat{\boldsymbol{\Delta}}_{\mathcal{N}})). \tag{39}$$

By rearranging (39), we obtain $(c-1)\mathcal{R}(\widehat{\boldsymbol{\Delta}}_{\mathcal{N}}) \leq (c+1)\mathcal{R}(\widehat{\boldsymbol{\Delta}}_{\mathcal{N}_\perp})$, which completes the proof. ∎

## Appendix B. Proof of Theorem 3

**Proof** We have

$$||\mathbf{X}\widehat{\boldsymbol{\Delta}} - \mathbf{Z}||_{2,1} - ||\mathbf{Z}||_{2,1} = \sum_{k=1}^{m}(||\mathbf{X}\widehat{\boldsymbol{\Delta}}_{*k} - \mathbf{Z}_{*k}||_2 - ||\mathbf{Z}_{*k}||_2)$$

$$= \sum_{k=1}^{m} \frac{||\mathbf{X}\widehat{\boldsymbol{\Delta}}_{*k}||_2^2 - 2(\mathbf{X}\widehat{\boldsymbol{\Delta}}_{*k})^T\mathbf{Z}_{*k}}{||\mathbf{X}\widehat{\boldsymbol{\Delta}}_{*k} - \mathbf{Z}_{*k}||_2 + ||\mathbf{Z}_{*k}||_2} \geq \sum_{k=1}^{m} \frac{||\mathbf{X}\widehat{\boldsymbol{\Delta}}_{*k}||_2^2}{||\mathbf{X}\widehat{\boldsymbol{\Delta}}_{*k}||_2 + 2||\mathbf{Z}_{*k}||_2} - 2\sum_{k=1}^{m} \frac{|\widehat{\boldsymbol{\Delta}}_{*k}^T\mathbf{X}^T\mathbf{Z}_{*k}|}{||\mathbf{Z}_{*k}||_2}. \tag{40}$$

Since $\mathbf{G}_{*k}^0 = \mathbf{X}^T\mathbf{Z}_{*k}/||\mathbf{Z}_{*k}||_2$, we have

$$\sum_{k=1}^{m} \frac{|\widehat{\boldsymbol{\Delta}}_{*k}^T\mathbf{X}^T\mathbf{Z}_{*k}|}{||\mathbf{Z}_{*k}||_2} = \sum_{k=1}^{m}|\widehat{\boldsymbol{\Delta}}_{*k}^T\mathbf{G}_{*k}^0| \leq \sum_{k=1}^{m}\sum_{j=1}^{d}|\widehat{\boldsymbol{\Delta}}_{jk}\mathbf{G}_{jk}^0| \leq \mathcal{R}^*(\mathbf{G}^0)\mathcal{R}(\widehat{\boldsymbol{\Delta}}), \tag{41}$$

where the last inequality follows from the Cauchy-Schwarz inequality. Recall that in the proof of Lemma 2, we already have (36) as follows,

$$||\mathbf{X}\widehat{\boldsymbol{\Delta}} - \mathbf{Z}||_{2,1} - ||\mathbf{Z}||_{2,1} \leq \lambda(\mathcal{R}(\widehat{\boldsymbol{\Delta}}_{\mathcal{N}_\perp}) - \mathcal{R}(\widehat{\boldsymbol{\Delta}}_{\mathcal{N}})). \tag{42}$$

Therefore by combining (42) with (40) and (41), we obtain

$$\sum_{k=1}^{m} \frac{||\mathbf{X}\widehat{\boldsymbol{\Delta}}_{*k}||_2^2}{||\mathbf{X}\widehat{\boldsymbol{\Delta}}_{*k}||_2 + 2||\mathbf{Z}_{*k}||_2} \leq \lambda\big(\mathcal{R}(\widehat{\boldsymbol{\Delta}}_{\mathcal{N}_\perp}) - \mathcal{R}(\widehat{\boldsymbol{\Delta}}_{\mathcal{N}})\big) + 2\mathcal{R}^*(\mathbf{G}^0)\mathcal{R}(\widehat{\boldsymbol{\Delta}})$$

$$\leq \lambda\,(1 + 2/c)\,\mathcal{R}(\widehat{\boldsymbol{\Delta}}_{\mathcal{N}_\perp}) + \lambda\,(2/c - 1)\,\mathcal{R}(\widehat{\boldsymbol{\Delta}}_{\mathcal{N}}) \leq \frac{2\lambda}{c-1}\mathcal{R}(\widehat{\boldsymbol{\Delta}}_{\mathcal{N}_\perp}), \tag{43}$$

where the second inequality comes from the assumption $\lambda \geq c\mathcal{R}^*(\mathbf{G}^0)$ and the triangle inequality $\mathcal{R}(\widehat{\boldsymbol{\Delta}}) \leq \mathcal{R}(\widehat{\boldsymbol{\Delta}}_{\mathcal{N}_\perp}) + \mathcal{R}(\widehat{\boldsymbol{\Delta}}_{\mathcal{N}})$, and the last inequality comes from (13) in Lemma 2. Meanwhile, by the triangle inequality, we also have

$$\sum_{k=1}^m \frac{||\mathbf{X}\widehat{\boldsymbol{\Delta}}_{*k}||_2^2}{||\mathbf{X}\widehat{\boldsymbol{\Delta}}_{*k}||_2 + 2||\mathbf{Z}_{*k}||_2} \geq \frac{\sum_{k=1}^m ||\mathbf{X}\widehat{\boldsymbol{\Delta}}_{*k}||_2^2}{||\mathbf{X}\widehat{\boldsymbol{\Delta}}||_{2,\infty} + 2||\mathbf{Z}||_{2,\infty}} \geq \frac{||\mathbf{X}\widehat{\boldsymbol{\Delta}}||_F^2}{||\mathbf{X}\widehat{\boldsymbol{\Delta}}||_F + 2||\mathbf{Z}||_{2,\infty}}, \tag{44}$$

where the last inequality comes from the fact $||\mathbf{X}\widehat{\boldsymbol{\Delta}}||_{2,\infty} \leq ||\mathbf{X}\widehat{\boldsymbol{\Delta}}||_F$. Combining (43) and (44), we obtain

$$\frac{||\mathbf{X}\widehat{\boldsymbol{\Delta}}||_F^2}{||\mathbf{X}\widehat{\boldsymbol{\Delta}}||_F + 2||\mathbf{Z}||_{2,\infty}} \leq \frac{2\lambda}{c-1}\mathcal{R}(\widehat{\boldsymbol{\Delta}}_{\mathcal{N}_\perp}) \leq \frac{2\lambda\Theta(\mathcal{N}_\perp, \mathcal{R})||\widehat{\boldsymbol{\Delta}}||_F}{c-1}, \tag{45}$$

where the last inequality comes from the definition of $\Theta(\mathcal{N}_\perp, \mathcal{R})$. By Assumption 1, we can rewrite (45) as

$$||\mathbf{X}\widehat{\boldsymbol{\Delta}}||_F^2 \leq \frac{2\lambda\Theta(\mathcal{N}_\perp, \mathcal{R})}{(c-1)\sqrt{n}\kappa}||\mathbf{X}\widehat{\boldsymbol{\Delta}}||_F^2 + \frac{4\lambda\Theta(\mathcal{N}_\perp, \mathcal{R})}{\sqrt{n}\kappa(c-1)}||\mathbf{Z}||_{2,\infty}||\mathbf{X}\widehat{\boldsymbol{\Delta}}||_F.$$

Given $2\Theta(\mathcal{N}_\perp, \mathcal{R}) \leq \delta(c-1)\sqrt{n}\kappa$ for some $\delta < 1$, we have

$$||\mathbf{X}\widehat{\boldsymbol{\Delta}}||_F \leq \frac{4\lambda\Theta(\mathcal{N}_\perp, \mathcal{R})}{\sqrt{n}\kappa(c-1)(1-\delta)}||\mathbf{Z}||_{2,\infty} \leq \frac{4\lambda\Theta(\mathcal{N}_\perp, \mathcal{R})\sigma_{\max}}{\sqrt{n}\kappa(c-1)(1-\delta)}||\mathbf{W}||_{2,\infty}. \tag{46}$$

By Assumption 1 again, we obtain

$$||\widehat{\boldsymbol{\Delta}}||_F \leq \frac{4\lambda\Theta(\mathcal{N}_\perp, \mathcal{R})\sigma_{\max}}{n\kappa^2(c-1)(1-\delta)}||\mathbf{W}||_{2,\infty}. \tag{47}$$

We proceed with the standard deviation estimation. By (36), we have

$$||\mathbf{Y} - \mathbf{X}\widehat{\mathbf{B}}||_{2,1} - ||\mathbf{Y} - \mathbf{X}\mathbf{B}^0||_{2,1} \leq \lambda\mathcal{R}(\widehat{\boldsymbol{\Delta}}_{\mathcal{N}_\perp}) - \lambda\mathcal{R}(\widehat{\boldsymbol{\Delta}}_{\mathcal{N}}) \leq \lambda\mathcal{R}(\widehat{\boldsymbol{\Delta}}_{\mathcal{N}_\perp}). \tag{48}$$

Combining (48) with a simple variant of Assumption 1

$$\kappa \leq \frac{||\mathbf{X}\widehat{\boldsymbol{\Delta}}||_F}{\sqrt{n}||\widehat{\boldsymbol{\Delta}}||_F} \leq \frac{||\mathbf{X}\widehat{\boldsymbol{\Delta}}||_F}{\sqrt{n}||\widehat{\boldsymbol{\Delta}}_{\mathcal{N}_\perp}||_F} \leq \frac{\Theta(\mathcal{N}_\perp, \mathcal{R})||\mathbf{X}\widehat{\boldsymbol{\Delta}}||_F}{\sqrt{n}\mathcal{R}(\widehat{\boldsymbol{\Delta}}_{\mathcal{N}_\perp})}, \tag{49}$$

we have

$$\sqrt{n}\left(\sum_{k=1}^m \widehat{\sigma}_k - \sum_{k=1}^m \sigma_k\right) \leq \frac{\lambda\Theta(\mathcal{N}_\perp, \mathcal{R})||\mathbf{X}\widehat{\boldsymbol{\Delta}}||_F}{\sqrt{n}\kappa} \leq \frac{4\lambda^2\Theta^2(\mathcal{N}_\perp, \mathcal{R})\sigma_{\max}}{n\kappa(c-1)(1-\delta)}||\mathbf{W}||_{2,\infty}, \tag{50}$$

where the last inequality comes from (46). By (37), (38), and Lemma 2, we have

$$||\mathbf{Y} - \mathbf{X}\widehat{\mathbf{B}}||_{2,1} - ||\mathbf{Y} - \mathbf{X}\mathbf{B}^0||_{2,1} \geq -\frac{\lambda}{c}(\mathcal{R}(\widehat{\boldsymbol{\Delta}}_{\mathcal{N}_\perp}) + \mathcal{R}(\widehat{\boldsymbol{\Delta}}_{\mathcal{N}})) \geq -\frac{2\lambda}{c-1}\mathcal{R}(\widehat{\boldsymbol{\Delta}}_{\mathcal{N}_\perp}). \tag{51}$$

By (49) again, we have

$$\sqrt{n}\left(\sum_{k=1}^m \widehat{\sigma}_k - \sum_{k=1}^m \sigma_k\right) \geq -\frac{8\lambda^2\Theta^2(\mathcal{N}_\perp, \mathcal{R})\sigma_{\max}}{n\kappa(c-1)^2(1-\delta)}||\mathbf{W}||_{2,\infty}. \tag{52}$$

Thus combining (50) and (52), we have

$$\frac{1}{m}\left|\sum_{k=1}^m \widehat{\sigma}_k - \sum_{k=1}^m \sigma_k\right| \leq \max\left\{1, \frac{2}{c-1}\right\}\frac{4\lambda^2\Theta^2(\mathcal{N}_\perp, \mathcal{R})\sigma_{\max}}{\sqrt{n}mn\kappa(c-1)(1-\delta)}||\mathbf{W}||_{2,\infty}. \tag{53}$$

∎

## Appendix C. Proof of Corollary 4

We need to introduce the following lemmas for our proof.

**Lemma 10** *Suppose that we have all entries of a random vector $\boldsymbol{v} = (v_1, ..., v_n)^T \in \mathbb{R}^n$ independently generated from the standard Gaussian distribution with mean 0 and variance 1. For any $c_0 \in (0, 1)$, we have*

$$\mathbb{P}\left(\left|\|\boldsymbol{v}\|_2^2 - n\right| \geq c_0 n\right) \leq 2\exp\left(-\frac{nc_0^2}{8}\right).$$

The proof of Lemma 10 is provided in Johnstone (2001), therefore omitted.

**Lemma 11** *Suppose that we have all entries of $\mathbf{W}$ independently generated from the standard Gaussian distribution with mean 0 and variance 1, then there exists some universal constant $c_1$ such that*

$$\mathbb{P}\left(\frac{\|\mathbf{X}^T\mathbf{W}\|_2}{\sqrt{n}} \leq \frac{2\|\mathbf{X}\|_2}{\sqrt{n}}(\sqrt{m} + \sqrt{d})\right) \geq 1 - 2\exp(-c_1(d+m)). \tag{54}$$

The proof of Lemma 11 is provided in Appendix E. Now we proceed to derive the refined error bound for the calibrated low rank regression estimator.

**Proof** Since we have all entries of $\mathbf{W}$ independently generated from $N(0, 1)$, then by Lemma 10, for any $c_0 \in (0, 1)$, we have

$$\mathbb{P}\left(\sqrt{(1-c_0)n} \leq \|\mathbf{W}_{*k}\|_2 \leq \sqrt{(1+c_0)n}\right) \geq 1 - 2\exp\left(-\frac{nc_0^2}{8}\right).$$

By taking the union bound over all $k = 1, ..., m$, we have

$$\mathbb{P}\left(\sqrt{(1-c_0)n} \leq \min_{1 \leq k \leq m} \|\mathbf{W}_{*k}\|_2 \leq \max_{1 \leq k \leq m} \|\mathbf{W}_{*k}\|_2 \leq \sqrt{(1+c_0)n}\right)$$
$$\geq 1 - 2m\exp\left(-\frac{nc_0^2}{8}\right). \tag{55}$$

Now conditioning on the event $\sqrt{(1-c_0)n} \leq \min_{1 \leq k \leq m} \|\mathbf{W}_{*k}\|_2$, we have

$$\mathcal{R}^*(\mathbf{G}^0) = \|\mathbf{G}^0\|_2 = \max_{\|\boldsymbol{v}\|_2 \leq 1} \sqrt{\sum_{k=1}^m \frac{(\boldsymbol{v}^T\mathbf{X}^T\mathbf{W}_{*k})^2}{\|\mathbf{W}_{*k}\|_2^2}}$$
$$\leq \max_{\|\boldsymbol{v}\|_2 \leq 1} \sqrt{\frac{\sum_{k=1}^m (\boldsymbol{v}^T\mathbf{X}^T\mathbf{W}_{*k})^2}{(1-c_0)n}} = \frac{\|\mathbf{X}^T\mathbf{W}\|_2}{\sqrt{(1-c_0)n}}. \tag{56}$$

By Lemma 11, there exists some universal positive constant $c_1$ such that we have

$$\mathbb{P}\left(\frac{\|\mathbf{X}^T\mathbf{W}\|_2}{\sqrt{(1-c_0)n}} \leq \frac{2\|\mathbf{X}\|_2(\sqrt{d}+\sqrt{m})}{\sqrt{n(1-c_0)}}\right) \geq 1 - 2\exp\left(-c_1(d+m)\right). \tag{57}$$

Given any matrix $\mathbf{A}$ in $\mathcal{N}_\perp$, $\mathbf{A}$ has at most rank $2r$ (See more details in Appendix B of Negahban and Wainwright (2011)). Then we have

$$\|\mathbf{A}\|_* = \sum_{j=1}^{2r} \psi_j(\mathbf{A}) \leq \sqrt{2r}\sqrt{\sum_{j=1}^{2r} \psi_j(\mathbf{A})^2} = \sqrt{2r}\|\mathbf{A}\|_{\mathrm{F}}.$$

Therefore we have $\Theta(\mathcal{N}_\perp, ||\cdot||_*) = \sqrt{2r}$. Theorem 3 requires

$$2\lambda\Theta(\mathcal{N}_\perp, \mathcal{R}) \leq \delta\kappa(c-1)\sqrt{n} \text{ for some } \delta < 1. \tag{58}$$

Thus if we take

$$\lambda = \frac{2c||\mathbf{X}||_2(\sqrt{m} + \sqrt{d})}{\sqrt{n}(1-c_0)},$$

then we need $n$ to be large enough

$$n \geq \frac{4\sqrt{2}c||\mathbf{X}||_2(\sqrt{rm} + \sqrt{rd})}{\delta\kappa(c-1)\sqrt{1-c_0}},$$

such that (58) can be secured. Then by combining (55), (56), (57), (47), and (53), we complete the proof. ∎

# Appendix D. Proof of Corollary 5

We need to introduce the following lemma for our proof.

**Lemma 12** *Suppose that we have all entries of $\mathbf{W}$ independently generated from the standard Gaussian distribution with mean 0 and variance 1, then we have*

$$\mathbb{P}\left(\max_{1 \leq j \leq d} \frac{1}{\sqrt{n}}||\mathbf{W}^T\mathbf{X}_{*j}||_q \leq 2\left(m^{1-1/p} + \sqrt{\log d}\right)\right) \geq 1 - \frac{2}{d},$$

*where $1/p + 1/q = 1$.*

The proof of Lemma 12 is provided in Appendix F. Now we proceed to derive the refined error bound for the joint sparsity setting.

**Proof** Recall that we already have (55),

$$\mathbb{P}\left(\sqrt{(1-c_0)n} \leq \min_{1 \leq k \leq m}||\mathbf{W}_{*k}||_2 \leq \max_{1 \leq k \leq m}||\mathbf{W}_{*k}||_2 \leq \sqrt{(1+c_0)n}\right)$$

$$\geq 1 - 2m\exp\left(-\frac{nc_0^2}{8}\right). \tag{59}$$

Now conditioning on the event $\sqrt{(1-c_0)n} \leq \min_{1 \leq k \leq m}||\mathbf{W}_{*k}||_2$, we have

$$\mathcal{R}^*(\mathbf{G}^0) = ||\mathbf{G}^0||_{\infty,q} = \max_{1 \leq j \leq d}\left(\sum_{k=1}^{n}\frac{(\mathbf{W}_{*k}^T\mathbf{X}_{*j})^q}{||\mathbf{W}_{*k}||_2^q}\right)^{1/q} \leq \frac{\max_{1 \leq j \leq d}||\mathbf{W}^T\mathbf{X}_{*j}||_q}{\min_{1 \leq k \leq m}||\mathbf{W}_{*k}||_2} \leq \frac{||\mathbf{X}^T\mathbf{W}||_{\infty,q}}{\sqrt{(1-c_0)n}}. \tag{60}$$

By Lemma 12, we have

$$\mathbb{P}\left(\frac{||\mathbf{X}^T\mathbf{W}||_{\infty,q}}{\sqrt{(1-c_0)n}} \leq \frac{2m^{1-1/p}}{\sqrt{(1-c_0)}} + \frac{2\sqrt{\log d}}{\sqrt{(1-c_0)}}\right) \geq 1 - \frac{2}{d}. \tag{61}$$

Given any matrix $\mathbf{A}$ in $\mathcal{N}_\perp$, $\mathbf{A}$ has at most $s$ nonzero rows. Then we have

$$||\mathbf{A}||_{1,p} = \sum_{\mathbf{A}_{j*} \neq \mathbf{0}}||\mathbf{A}_{j*}||_p \leq \sum_{\mathbf{A}_{j*} \neq \mathbf{0}}||\mathbf{A}_{j*}||_2 \leq \sqrt{s}\sqrt{\sum_{\mathbf{A}_{j*} \neq \mathbf{0}}||\mathbf{A}_{j*}||_2^2} = \sqrt{s}||\mathbf{A}||_F.$$

Therefore we have $\Theta(\mathcal{N}_\perp, ||\cdot||_{1,p}) = \sqrt{s}$ for any $2 \le p \le \infty$. Theorem 3 requires

$$2\lambda\Theta(\mathcal{N}_\perp, \mathcal{R}) \le \delta\kappa(c-1)\sqrt{n} \text{ for some } \delta < 1. \tag{62}$$

Thus if we take

$$\lambda = \frac{2c(m^{1-1/p} + \sqrt{\log d})}{\sqrt{1-c_0}},$$

then we need $n$ to be large enough

$$\sqrt{n} \ge \frac{4c\sqrt{s}(m^{1-1/p} + \sqrt{\log d})}{\delta\kappa(c-1)\sqrt{1-c_0}},$$

such that (62) can be secured. Then by combining (59), (60), (61), (47), and (53), we complete the proof.

∎

# Appendix E. Proof of Lemma 11

**Proof** Since $\mathbf{W}$ has all its entries independently generated from the standard Gaussian distribution with mean 0 and variance 1, then all $\mathbf{X}^T\mathbf{W}_{*k}/\sqrt{n}$'s are essentially independently generated from a multivariate Gaussian distribution with mean $\mathbf{0}$ and covariance matrix $\mathbf{X}^T\mathbf{X}/n$.

Thus by Corollary 5.50 in Vershynin (2010) on the singular values of Gaussian random matrices (Davidson and Szarek, 2001), we know that there exists a universal positive constant $c_1$ such that

$$\mathbb{P}\left(\frac{||\mathbf{X}^T\mathbf{W}||_2}{\sqrt{n}} \le \frac{2||\mathbf{X}||_2}{\sqrt{n}}(\sqrt{m} + \sqrt{d})\right) \ge 1 - 2\exp(-c_1(d+m)), \tag{63}$$

which completes the proof.

∎

# Appendix F. Proof of Lemma 12

**Proof** We adopt the similar proof strategy in Negahban et al. (2012), and begin our proof by establishing the tail bound of $||\mathbf{W}^T\mathbf{X}_{*j}||_q/\sqrt{n}$.

**Deviation above the mean**: Given any pair of $\mathbf{W}, \widetilde{\mathbf{W}} \in \mathbb{R}^{n \times m}$ and $1/q + 1/p = 1$, we have

$$\left|\frac{1}{\sqrt{n}}||\mathbf{W}^T\mathbf{X}_{*j}||_q - \frac{1}{\sqrt{n}}||\widetilde{\mathbf{W}}^T\mathbf{X}_{*j}||_q\right| \le \frac{1}{\sqrt{n}}||(\mathbf{W} - \widetilde{\mathbf{W}})^T\mathbf{X}_{*j}||_q$$
$$= \frac{1}{\sqrt{n}}\max_{||\boldsymbol{\theta}||_p \le 1}\langle\boldsymbol{\theta}, (\mathbf{W} - \widetilde{\mathbf{W}})^T\mathbf{X}_{*j}\rangle. \tag{64}$$

By the Cauchy-Schwartz inequality, we have

$$\frac{1}{\sqrt{n}}\max_{||\boldsymbol{\theta}||_p \le 1}\langle\boldsymbol{\theta}\mathbf{X}_{*j}^T, \mathbf{W} - \widetilde{\mathbf{W}}\rangle \le \frac{||\mathbf{W} - \widetilde{\mathbf{W}}||_F}{\sqrt{n}}\max_{||\boldsymbol{\theta}||_p \le 1}||\boldsymbol{\theta}\mathbf{X}_{*j}^T||_F. \tag{65}$$

Since $\boldsymbol{\theta}\mathbf{X}_{*j}^T$ is a rank one matrix, its singular value decomposition is

$$\boldsymbol{\theta}\mathbf{X}_{*j}^T = ||\boldsymbol{\theta}||_2||\mathbf{X}_{*j}|| \cdot \frac{\boldsymbol{\theta}}{||\boldsymbol{\theta}||_2} \cdot \frac{\mathbf{X}_{*j}^T}{||\mathbf{X}_{*j}||_2}.$$

Consequently, we have

$$\frac{1}{\sqrt{n}} \max_{||\boldsymbol{\theta}||_p \leq 1} ||\boldsymbol{\theta} \mathbf{X}_{*j}^T||_F = \frac{||\mathbf{X}_{*j}||_2}{\sqrt{n}} \max_{||\boldsymbol{\theta}||_p \leq 1} ||\boldsymbol{\theta}||_2 \overset{\text{(i)}}{\leq} \frac{m^{1/2-1/p}||\mathbf{X}_{*j}||_2}{\sqrt{n}} \overset{\text{(ii)}}{\leq} 1. \tag{66}$$

where (i) comes from $||\boldsymbol{\theta}||_2 \leq m^{1/2-1/p}||\boldsymbol{\theta}||_p$, and (ii) comes from the column normalization condition (21). Combining (64), (65), and (66), we obtain

$$\left| \frac{1}{\sqrt{n}} ||\mathbf{W}^T \mathbf{X}_{*j}||_q - \frac{1}{\sqrt{n}} ||\widetilde{\mathbf{W}}^T \mathbf{X}_{*j}||_q \right| \leq ||\mathbf{W} - \widetilde{\mathbf{W}}||_F. \tag{67}$$

which implies that $||\mathbf{W}^T \mathbf{X}_{*j}||_q / \sqrt{n}$ is a Lipschitz continuous function of $\mathbf{W}$ with a Lipschitz constant as 1. By the Gaussian concentration of measure for Lipschitz functions (Ledoux and Talagrand, 2011), we have

$$\mathbb{P}\left( \frac{1}{\sqrt{n}} ||\mathbf{W}^T \mathbf{X}_{*j}||_q \geq \mathbb{E} \frac{1}{\sqrt{n}} ||\mathbf{W}^T \mathbf{X}_{*j}||_q + \xi \right) \leq 2 \exp\left( -\frac{\xi^2}{2} \right). \tag{68}$$

**Upper bound of the mean**: Given any $\boldsymbol{\beta} \in \mathbb{R}^m$, we define a zero mean Gaussian random variable $J_{\boldsymbol{\beta}} = \boldsymbol{\beta}^T \mathbf{W}^T \mathbf{X}_{*j} / \sqrt{n}$, and note that we have $\frac{1}{\sqrt{n}} ||\mathbf{W}^T \mathbf{X}_{*j}||_q = \max_{||\boldsymbol{\beta}||_p=1} J_{\boldsymbol{\beta}}$. Thus given any two vectors $||\boldsymbol{\beta}||_p \leq 1$ and $||\boldsymbol{\beta}'||_p \leq 1$, we have

$$\mathbb{E}(J_{\boldsymbol{\beta}} - J_{\boldsymbol{\beta}'})^2 = \frac{1}{n} ||\mathbf{X}_{*j}||_2^2 ||\boldsymbol{\beta} - \boldsymbol{\beta}'||_2^2 \leq ||\boldsymbol{\beta} - \boldsymbol{\beta}'||_2^2,$$

where the last inequality comes from (21) and $m^{1-1/p} \geq 1$.

Then we define another Gaussian random variable $K_{\boldsymbol{\beta}} = \boldsymbol{\beta}^T \boldsymbol{\omega}$, where $\boldsymbol{\omega} = (\omega_1, ..., \omega_m)^T \sim N(\mathbf{0}, \mathbf{I}_m)$ is standard Gaussian. By construction, for any pair $\boldsymbol{\beta}, \boldsymbol{\beta}' \in \mathbb{R}^m$, we have

$$\mathbb{E}[(K_{\boldsymbol{\beta}} - K_{\boldsymbol{\beta}'})^2] = ||\boldsymbol{\beta} - \boldsymbol{\beta}'||_2^2 \geq \mathbb{E}(J_{\boldsymbol{\beta}} - J_{\boldsymbol{\beta}'})^2.$$

Thus by the Sudakov-Fernique comparison principle (Ledoux and Talagrand, 2011), we have

$$\mathbb{E} \frac{1}{\sqrt{n}} ||\mathbf{W}^T \mathbf{X}_{*j}||_q = \mathbb{E} \max_{||\boldsymbol{\beta}||_p=1} J_{\boldsymbol{\beta}} \leq \mathbb{E} \max_{||\boldsymbol{\beta}||_p=1} K_{\boldsymbol{\beta}}.$$

By definition of $K_{\boldsymbol{\beta}}$, we have

$$\mathbb{E} \max_{||\boldsymbol{\beta}||_p=1} K_{\boldsymbol{\beta}} = \mathbb{E} ||\boldsymbol{\omega}||_q \leq m^{1/q} (\mathbb{E}|\boldsymbol{\omega}_1|^q)^{1/q}, \tag{69}$$

where the last inequality comes from Jensen's inequality and the fact that $|\boldsymbol{\omega}_1|^{1/q}$ is a concave function of $\boldsymbol{\omega}_1$ for $q \in [1, 2]$. Eventually, by Hölder inequality, we obtain

$$(\mathbb{E}|\boldsymbol{\omega}_1|^q)^{1/q} \leq \sqrt{\mathbb{E}\omega_1^2} = 1. \tag{70}$$

Combing (69) and (70), we obtain

$$\mathbb{E} \max_{||\boldsymbol{\beta}||_p=1} K_{\boldsymbol{\beta}} \leq m^{1-1/p} \leq 2m^{1-1/p}. \tag{71}$$

Then combing (68) and (71), we have

$$\mathbb{P}\left( \frac{1}{\sqrt{n}} ||\mathbf{W}^T \mathbf{X}_{*j}||_q \geq 2m^{1-1/p} + \xi \right) \leq 2 \exp\left( -\frac{\xi^2}{2} \right).$$

Taking the union bound over $j = 1, ..., d$ and let $\xi = 2\sqrt{\log d}$, we have

$$\mathbb{P}\left( \frac{1}{\sqrt{n}} ||\mathbf{X}^T \mathbf{W}||_{\infty,q} \geq 2m^{1-1/p} + 2\sqrt{\log d} \right) \leq \frac{2}{d}.$$

This finishes the proof. ∎